\documentclass[runningheads]{llncs}

\usepackage{eccv}

\usepackage{eccvabbrv}

\usepackage{graphicx}
\usepackage{booktabs}
\usepackage{multirow}
\usepackage[table,xcdraw]{xcolor}
\definecolor{no1}{HTML}{FFB2B2}  % 定义浅红色 (no1)
\definecolor{no2}{HTML}{FFD9B2}  % 定义浅橙色 (no2)
\usepackage{caption}
\usepackage{booktabs} % 绘制三线表 (\toprule, \midrule, \bottomrule)
\usepackage{pifont}   % 绘制对钩和叉号 (\ding{51}, \ding{55})
\usepackage{lmodern}

\usepackage[accsupp]{axessibility}  % Improves PDF readability for those with disabilities.

\usepackage{hyperref}

\usepackage{orcidlink}

\makeatletter

\renewcommand{\footnoterule}{%
  \kern -3pt
  \hrule width .25\columnwidth height 0.2pt
  \kern 2.6pt
}

\newcommand{\titlefootnotes}[1]{%
  \begingroup
  \renewcommand{\thefootnote}{}%
  \footnotetext[0]{#1}%
  \endgroup
}

\makeatother

\begin{document}

% ---------------------------------------------------------------
% TODO REVIEW: Replace with your title
\title{Robust and Efficient Monocular 3D Gaussian SLAM for Kilometer-Scale Outdoor Scenes} 

% TODO REVIEW: If the paper title is too long for the running head, you can set
% an abbreviated paper title here. If not, comment out.
% \titlerunning{Abbreviated paper title}
\titlerunning{KiloGS-SLAM}

% TODO FINAL: Replace with your author list. 
% Include the authors' OCRID for the camera-ready version, if at all possible.
% \author{Sicheng Yu\inst{1}\orcidlink{0000-1111-2222-3333} \and

\author{
Sicheng Yu\textsuperscript{*} \and
Dongxu Shen\textsuperscript{*} \and
Beizhen Zhao\and
Guanzhi Ding\and
Hao Wang\textsuperscript{\textdagger}
}

% TODO FINAL: Replace with an abbreviated list of authors.
% \authorrunning{F.~Author et al.}
\authorrunning{S.~Yu et al.}
% First names are abbreviated in the running head.
% If there are more than two authors, 'et al.' is used.

% TODO FINAL: Replace with your institution list.
\institute{
% The Hong Kong University of Science and Technology (Guangzhou), Guangzhou, China
% HKUST(GZ), Guangzhou, China
The Hong Kong University of Science and Technology (Guangzhou), China
}
% \institute{Princeton University, Princeton NJ 08544, USA \and
% Springer Heidelberg, Tiergartenstr.~17, 69121 Heidelberg, Germany
% \email{lncs@springer.com}\\
% \url{http://www.springer.com/gp/computer-science/lncs} \and
% ABC Institute, Rupert-Karls-University Heidelberg, Heidelberg, Germany\\
% \email{\{abc,lncs\}@uni-heidelberg.de}}

\maketitle

\titlefootnotes{
\textsuperscript{*}Equal contribution, \quad
\textsuperscript{$\dagger$}Corresponding author: \texttt{haowang@hkust-gz.edu.cn}}

% \begingroup
% \renewcommand{\thefootnote}{\fnsymbol{footnote}}
% \footnotetext[1]{Equal contribution.}
% \footnotetext[2]{Corresponding author: Hao Wang, \texttt{haowang@hkust-gz.edu.cn}.}
% \endgroup

\begin{center}
    \captionsetup{type=figure}
    \includegraphics[width=1.0\textwidth]{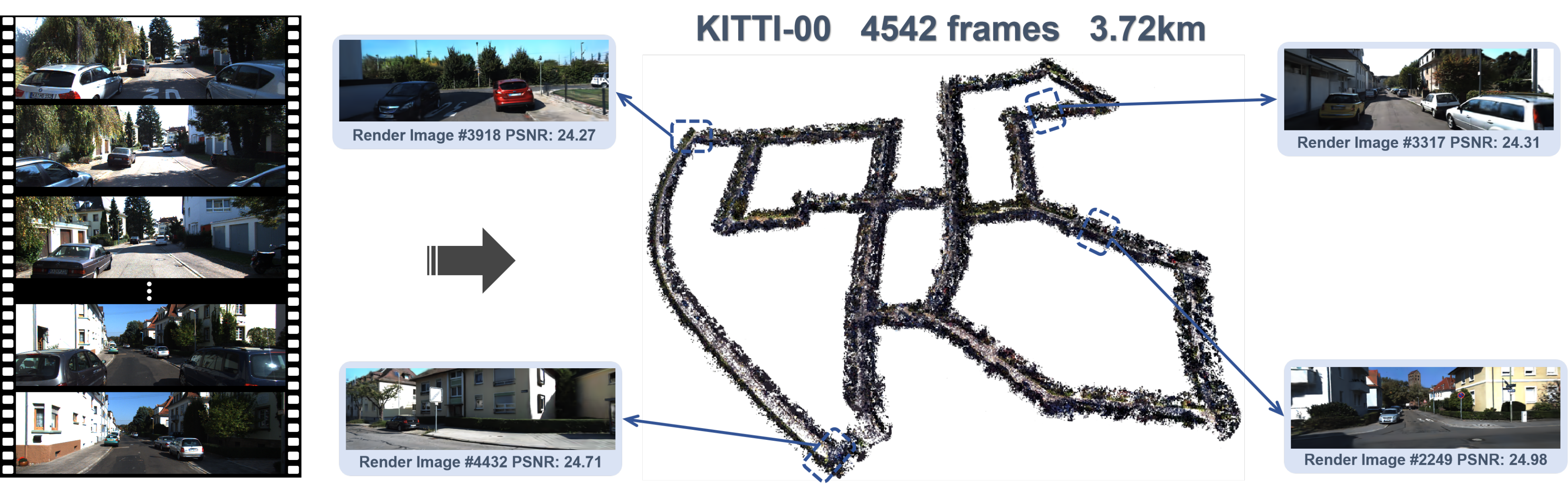}
    % \includegraphics[width=1.0\textwidth]{figs/排版占位.pdf}
    % \vspace{-10pt}
    \captionof{figure}{\textbf{KiloGS-SLAM} achieves robust pose tracking and precise scene reconstruction in kilometer-scale outdoor environments using only monocular RGB input, enabling high-fidelity novel view synthesis.}
    \label{teaser}
    % \vspace{-10pt}
\end{center}

\begin{abstract}

    Scaling monocular 3D Gaussian Splatting (3DGS) SLAM to kilometer-level outdoor environments poses two tightly coupled challenges: fragile long-term pose tracking and excessive memory overhead during large-scale mapping. 
    In this paper, we propose KiloGS-SLAM, a highly efficient and robust monocular 3DGS-SLAM system that jointly addresses both bottlenecks.
    Since high-fidelity scene reconstruction fundamentally relies on drift-free camera poses, we first introduce a motion-adaptive hybrid tracking module. This module features a condition-triggered three-tier solving pipeline. It dynamically switches between Essential matrix and PnP models to handle geometric degeneracies. An on-demand foundation model can also be activated to rescue the trajectory from catastrophic drift.
    To ensure the system can sustain these long trajectories without memory exhaustion, we subsequently design a lifecycle-managed Gaussian mapping strategy. 
    By integrating probabilistic initialization with chunk-based multi-view densification and pruning, this full-pipeline optimization effectively reduces primitive redundancy while preserving high-frequency details.
    Together, the robust tracking guarantees the geometric foundation required for accurate mapping, while the memory-efficient lifecycle-managed mapping enables large-scale operation.
    Extensive experiments across three challenging outdoor datasets demonstrate that our approach achieves state-of-the-art tracking accuracy and rendering quality, successfully scaling to sequences of over 10,000 frames on a single GPU. 
    Project Page: \url{https://3dagentworld.github.io/KiloGS-SLAM/}
  
  \keywords{Monocular SLAM \and 3D Gaussian Splatting \and Large-Scale Reconstruction \and Outdoor Environments}
\end{abstract}

% \clearpage

%%======================================1-Intro============================================

\section{Introduction}
\label{sec:intro}

While 3D Gaussian Splatting (3DGS) \cite{kerbl2023-3dgs,lu2024scaffold-gs,zhao2026mmgs} has emerged as a powerful representation for SLAM \cite{wang2026towards,keetha2024splatam,yan2024gs-slam}, offering real-time and photorealistic rendering, scaling it to kilometer-level outdoor scenes remains challenging. Although multi-sensor setups (e.g., LiDAR \cite{wu2024mmgaussian,hong2024LIVGaussMap,deng2025vpgs-slam} or IMU \cite{wu2025vings-mono}) can provide reliable geometric initialization, monocular RGB-only approaches remain highly desirable for their minimal hardware dependency. However, current monocular 3DGS-SLAM systems are fundamentally bottlenecked by two tightly coupled limitations: \textbf{fragile pose estimators} in complex environments and \textbf{excessive memory consumption} over long trajectories.

First, robust long-term tracking in outdoor environments is difficult. Traditional multi-view geometric solvers \cite{campos2021orb-slam3,teed2021droid-slam,teed2023dpvo} frequently collapse when confronting planar degeneracies (e.g., flat roads), pure rotations, or dynamic occlusions. Decoupled 3DGS-SLAM systems \cite{sandstrom2025splat-slam,deng2025gigaslam} that inherit these geometric frontends naturally suffer from the same vulnerability, often resulting in irreversible cumulative drift. While some recent approaches \cite{murai2025mast3r-slam,maggio2025vggt-slam,deng2025vggt-long,cheng2026longstream} attempt to replace geometric solvers with 3D foundation models, these inference-based methods are computationally heavy and lack rigorous geometric verification, limiting their continuous tracking precision. 

Second, the large scale and extreme depth variations of outdoor scenes make Gaussian mapping highly memory-intensive. Naive initialization and densification strategies \cite{matsuki2024monogs} inevitably cause an unconstrained growing of Gaussian primitives. Existing systems either introduce heavy hierarchical structures that fail to address the root cause of floater accumulation \cite{deng2025gigaslam}, or employ local heuristic pruning \cite{wu2025vings-mono} that ignores multi-view consistency, often erroneously deleting valid scene structures. 

To address these issues, we propose \textbf{KiloGS-SLAM}, a RGB-only monocular 3DGS-SLAM system designed specifically for kilometer-scale outdoor environments (Fig. \ref{teaser}). We argue that successfully scaling 3DGS-SLAM requires a systematic design: the pose estimator must provide a robust geometric foundation under varying conditions, while the mapping process must effectively minimize redundancy to sustain continuous operation.

To ensure stable tracking, we introduce a motion-adaptive hybrid tracking module. Instead of relying on a single solver, we design a condition-triggered, three-tier solving pipeline. Under normal conditions, it utilizes an efficient Essential matrix model. When standard geometry degrades (e.g., flat-ground scenes), it dynamically switches to a Perspective-n-Point (PnP) model. Crucially, to prevent severe drift, the system continuously verifies tracking consistency against motion priors. Any detected anomaly triggers an on-demand foundation model \cite{leroy2024mast3r} activation. By using foundation models as a supplementary model rather than a primary pose estimator, we achieve the speed and precision of geometric solvers, backed by the robustness of deep priors.

Given these reliable poses, we tackle the memory bottleneck by proposing a lifecycle-managed Gaussian mapping strategy. Rather than allowing Gaussians to grow randomly, this full-pipeline approach utilizes depth gradients and the Difference of Gaussians (DoG) for probabilistic initialization. As the map evolves, we continuously evaluate multi-view photometric errors within local chunks to guide densification and pruning. This lifecycle-managed optimization effectively suppresses redundant floaters while preserving high-frequency textural details, ensuring an optimal and lightweight primitive distribution.

We evaluate our system on the Waymo \cite{sun2020waymo}, KITTI \cite{geiger2013kitti}, and KITTI-360 \cite{liao2022kitti-360} datasets, encompassing nearly 30 sequences ranging from 200 to over 13,000 frames. Experimental results demonstrate that our approach achieves SOTA performance in both tracking and rendering. By achieving an optimal balance between visual fidelity and computational constraints (Fig. \ref{result:balance}), our system scales seamlessly to ultra-long scenes. Our main contributions are:

\begin{figure*}[t]
\centerline{\includegraphics[width=1.0\textwidth]{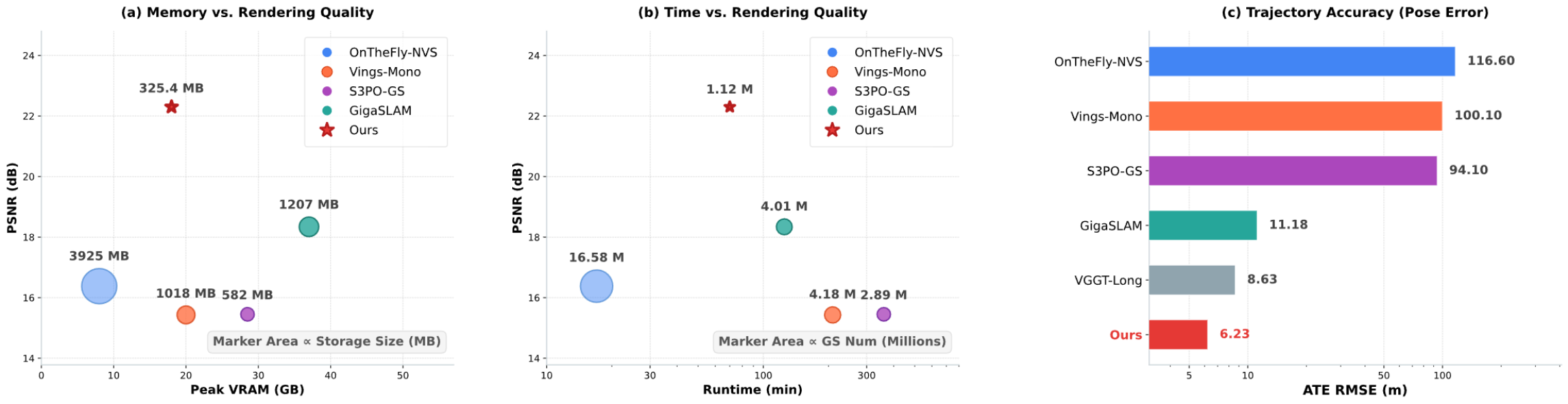}}
\vspace{-2pt}
\caption{\textbf{Performance comparison on KITTI-00.} Our method strikes the optimal balance between rendering quality, runtime, and memory overhead, while achieving the lowest camera tracking error.} 
\vspace{-5pt}
\label{result:balance}
\end{figure*}

\begin{enumerate}
    \item We propose a motion-adaptive hybrid tracking module featuring a condition-triggered three-tier pipeline. By dynamically switching between geometric solvers and an on-demand foundation model activation, it ensures robust tracking in kilometer-level outdoor sequences.
    \item We design a lifecycle-managed Gaussian mapping strategy. By integrating probabilistic initialization and chunk-based multi-view pruning and densification into a continuous optimization pipeline, it minimizes redundancy and memory overhead while maintaining high reconstruction quality.
    \item We achieve SOTA tracking and rendering performance across three demanding datasets, successfully scaling to continuous sequences exceeding 10,000 frames on a single GPU.
\end{enumerate}

%%======================================2-Related Work============================================
\section{Related Work}
%%%%%%%%%%%%%%%%%%%%%%%%%%%%%%%%%%%%%%  VO & SLAM  %%%%%%%%%%%%%%%%%%%%%%%%%%%
\subsection{Visual Odometry \& SLAM in Large-Scale Environments}
\label{sec:related_vo}

Monocular SLAM in large-scale outdoor environments remains a challenging task due to environmental complexities and long-term drift. 
% Early methods like LSD-SLAM \cite{engel2014lsd-slam} pioneered large-scale mapping, while the ORB-SLAM series \cite{mur2017orb-slam2,campos2021orb-slam3} stands as the gold standard for sparse feature-based systems. However, these methods are fragile in low-texture or highly dynamic driving scenarios. 
Early methods like LSD-SLAM \cite{engel2014lsd-slam} and ORB-SLAM series \cite{mur2017orb-slam2,campos2021orb-slam3} pioneered large-scale tracking but remain fragile in low-texture or dynamic outdoor scenarios.
% With the advent of deep learning, DF-VO \cite{zhan2021df-vo} utilized deep predictions to mitigate scale drift, and RoMeO \cite{cheng2024romeo} further exploited metric depth priors for robust scale recovery. 
% DROID-SLAM \cite{teed2021droid-slam} and DPVO \cite{teed2023dpvo} significantly enhanced robustness by leveraging recurrent optical flow and dense bundle adjustment.
With the advent of deep learning, DF-VO \cite{zhan2021df-vo} and RoMeO \cite{cheng2024romeo} leveraged depth priors to enhance tracking stability, while DROID-SLAM \cite{teed2021droid-slam} and DPVO \cite{teed2023dpvo} significantly enhanced robustness via recurrent optical flow and dense bundle adjustment.
DPV-SLAM \cite{lipson2024dpv-slam} further mitigated long-term drift by incorporating appearance-geometric loop closure. 
% Despite these advances, such approaches remain bound by geometric constraints. In degenerate outdoor scenarios—characterized by flat roads, pure rotations, or dynamic occlusions—reliable geometric cues become scarce, rendering these systems prone to instability.
Despite these advances, such methods rely heavily on multi-view geometric constraints, which frequently fail in complex outdoor scenarios characterized by planar degeneracies (e.g., flat roads), pure rotational motions, and dynamic occlusions, rendering these systems prone to tracking instability.

% Recent 3D Foundation Models (\eg, MASt3R \cite{leroy2024mast3r}, VGGT \cite{wang2025vggt}) attempt to transcend these limits by directly regressing geometry.
Recent 3D foundation models, like DUSt3R \cite{wang2024dust3r}, MASt3R \cite{leroy2024mast3r}, and VGGT \cite{wang2025vggt}, attempt to transcend these limitations by directly regressing 3D geometry and pose from large-scale pre-training.
% Recent 3D Foundation Models attempt to transcend these geometric limitations. DUSt3R \cite{wang2024dust3r}, MASt3R \cite{leroy2024mast3r}, and VGGT \cite{wang2025vggt} demonstrate the potential to directly regress 3D geometry and pose from large-scale pre-training. 
To adapt these models to long sequences, recent approaches employ a ``chunk-and-align'' strategy \cite{murai2025mast3r-slam,maggio2025vggt-slam,deng2025vggt-long,cheng2025unposed,cheng2025reggs}. However, they are limited by the local consistency of the base model, often suffering from temporal discontinuities and geometric misalignments at sub-map boundaries. 
% Most recently, LongStream \cite{cheng2026longstream} achieved kilometer-scale streaming inference via a gauge-decoupled architecture. Yet, as a end-to-end inference model, it lacks the rigorous geometric verification and iterative optimization inherent to SLAM systems; consequently, its tracking accuracy falls short of optimization-based frameworks. 

This motivates our design to integrate foundation models not as pose solvers, but as an on-demand activation mechanism for unreliable geometric estimation.
% This limitation motivates our design: instead of replacing the geometric solver entirely, we integrate the 3D foundation model as a complementary \textit{fallback mechanism}, activated only when geometric estimation becomes unreliable.

%%%%%%%%%%%%%%%%%%%%%%%%%%%%%%%%%%%%%%  GS-SLAM  %%%%%%%%%%%%%%%%%%%%%%%%%%%
\subsection{Gaussian Splatting SLAM}
\label{sec:related_gs}
The emergence of 3D Gaussian Splatting (3DGS) \cite{kerbl2023-3dgs,lu2024scaffold-gs,zhao2025wavelet-gs,zhao2026skewgs} has revolutionized SLAM with its real-time and high-fidelity rendering. 
Early 3DGS-based systems integrated explicit Gaussian representations for joint tracking and mapping \cite{matsuki2024monogs,keetha2024splatam,yan2024gs-slam,yugay2023gaussian-slam,hu2024cg-slam} or loosely coupled them with Visual Odometry \cite{huang2024photo-slam,sandstrom2025splat-slam,zhu2024mgs-slam,homeyer2025droid-splat}. 
While effective in indoor environments, these methods struggle in outdoor sequences, bottlenecked by fragile frontend robustness and excessive memory overhead.

To address these scalability issues, sensor-fused approaches \cite{wu2024mmgaussian,hong2024LIVGaussMap,deng2025vpgs-slam} leverage LiDAR for geometric initialization, ensuring structural accuracy but restricting deployment due to hardware dependency. 
VINGS-Mono \cite{wu2025vings-mono} introduces a visual-inertial framework with score-based pruning and concurrent CPU-GPU scheduling to mitigate memory pressure; however, its stability relies on IMU.
%rather than pure visual solving.

In the more challenging monocular setting, OpenGS-SLAM \cite{yu2025opengs-slam} and S3PO-GS \cite{cheng2025s3po-gs} incorporate geometric priors and scale correction.
While effective for street-level scenes, they lack efficient data structures for kilometer-scale sequences, causing memory exhaustion and drift. 
OnTheFly-NVS \cite{meuleman2025on-the-fly} achieves rapid reconstruction but suffers from significant accuracy degradation in long sequences.
Most related to our work is GigaSLAM \cite{deng2025gigaslam}, the first pure visual kilometer-scale 3DGS-SLAM system using a Hierarchical Sparse Voxel Map. Yet, it suffers from Gaussian redundancy and relies on heavy post-optimization. 

To bridge this gap, we propose a lifecycle-managed Gaussian mapping strategy, which integrates probabilistic Gaussian initialization and adaptive management based on multi-view consistency, ensuring efficient, high-fidelity mapping.
%%======================================3-Method============================================

\begin{figure*}
% \centerline{\includegraphics[width=1.0\textwidth, trim=0cm 8cm 11cm 0cm, clip]{figs/排版占位.pdf}}
% \centerline{\includegraphics[width=1.0\textwidth]{figs/排版占位.pdf}}
\centerline{\includegraphics[width=1.0\textwidth]{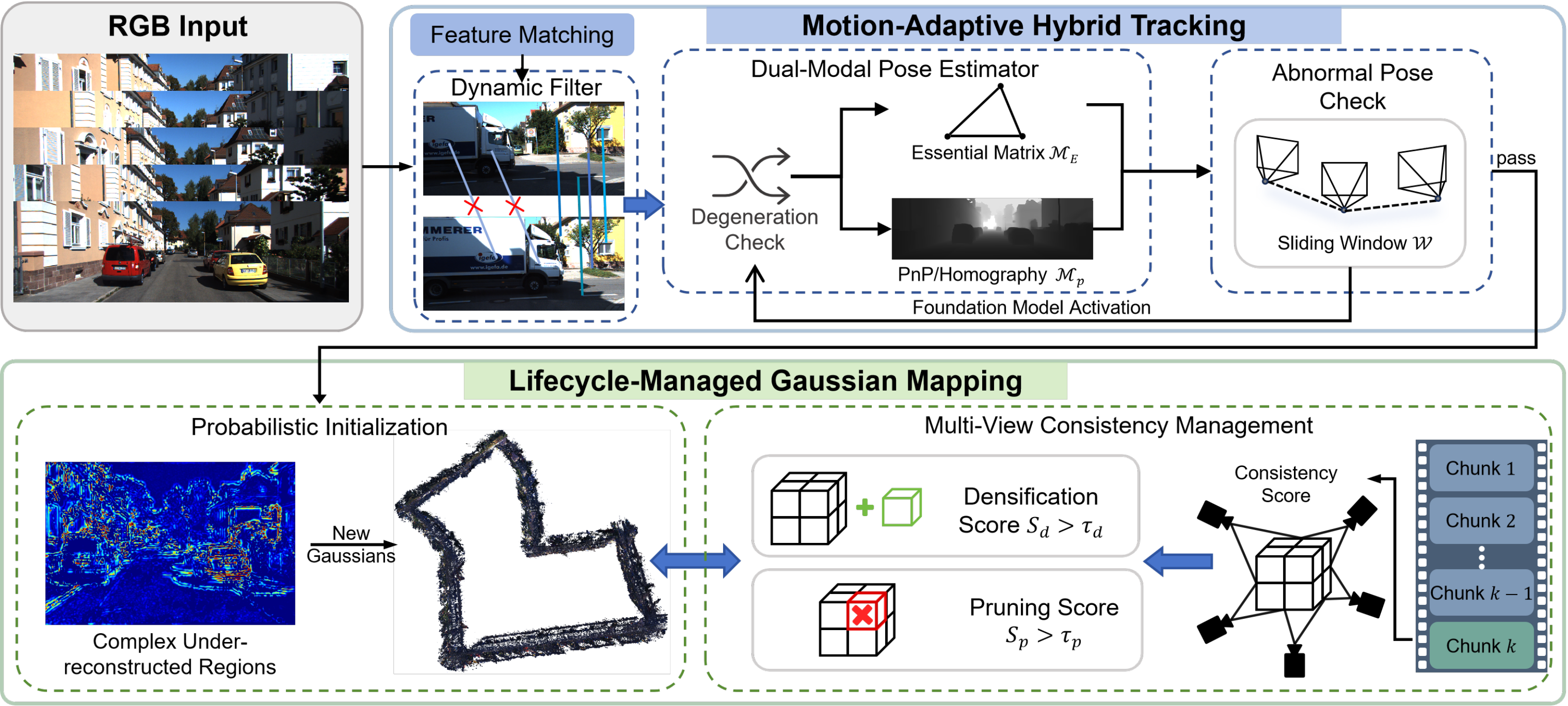}}
% \vspace{-5pt}
% \caption{\textbf{Framework of Ours.} \textbf{Tracking:} Input RGB frames undergo sparse matching and dynamic filtering. Poses are then estimated using an Adaptive Dual-Modal Solver and verified against a kinematic prior, which triggers a foundation model fallback upon failure. \textbf{Mapping:} Built upon a hierarchical voxelized representation, new Gaussians are initialized on-demand guided by DoG and depth gradients. Primitives are continually managed via chunk-based multi-view consistency scores.}
\caption{\textbf{Framework of KiloGS-SLAM.} Input RGB frames first undergo sparse matching and dynamic filtering once entering the tracking module. A dual-modal pose estimator dynamically switches between Essential and PnP matrices to handle degeneracies. The estimation is verified against a sliding-window motion prior, triggering an on-demand foundation model dense matching upon failure. Valid poses proceed to the Mapping module. New Gaussians are probabilistically initialized strictly in complex, under-reconstructed regions. During map optimization, chunk-based multi-view consistency scores are evaluated to guide continuous densification and pruning.}
% \vspace{-5pt}
\label{fig:framework}
\end{figure*}

\section{Method}
As illustrated in Fig.~\ref{fig:framework}, KiloGS-SLAM sequentially processes incoming RGB frames, estimates camera poses with Motion-adaptive Hybrid Tracking, and performs Lifecycle-managed Gaussian Mapping over keyframes.
%%%%%%%%%%%%%%%%%%%%%%%%%%%%%%%%%%%%%%  Frontend  %%%%%%%%%%%%%%%%%%%%%%%%%%%
\subsection{Motion-Adaptive Hybrid Tracking}
\label{sec:frontend}

The core of long-sequence pose tracking lies in consistent robustness. In kilometer-scale trajectories, tracking errors in merely a few frames can lead to irreversible cumulative drift. Fundamentally, the robustness of monocular visual odometry depends on two factors: the numerical stability of the pose solver and the reliability of feature correspondences. Complex outdoor environments severely challenge both. Large depth variations and planar degeneracies often render a single geometric solver ill-conditioned. Simultaneously, dramatic illumination changes, textureless regions, and high-frequency structural clutter (\eg, foliage) frequently induce massive spatial aliasing, leading to erroneous matches.

To tackle this dual challenge, we propose an adaptive hybrid tracking module. First, we utilize sparse optical flow statistics to filter out dynamic interference. The purified input is then fed into a condition-triggered, three-tier solving pipeline to combat varying degrees of geometric degeneracies. Crucially, to prevent severe mismatches from inducing physically implausible trajectory drift, we introduce a sliding-window motion consistency check. If an anomaly is detected, the system triggers a foundation model-based dense matching activation, rescuing the trajectory from severe visual degradation.

\subsubsection{Dual-Modal Pose Estimator}
\label{sec:dual_solver}
Relying on a single pose solver is fragile in complex outdoor scenarios due to common geometric degeneracies. Therefore, we introduce a dual-modal pose estimator as the primary component of our tracking module. Specifically, after establishing frame-to-frame correspondences via lightweight feature matching \cite{tyszkiewicz2020disk}, we adaptively switch between an epipolar-based Essential matrix model ($\mathcal{M}_E$) \cite{nister2004epipolar} and a Perspective-n-Point/Homography model ($\mathcal{M}_P$) \cite{lepetit2009epnp} for pose estimation. This switching is inspired by the Geometric Robust Information Criterion (GRIC) \cite{zhan2021df-vo} score, effectively handling geometric degeneracies like pure rotations or flat-ground scenes. 

\subsubsection{Motion Consistency Verification}
\label{sec:kinematic_check}
To detect tracking anomalies caused by massive mismatches, we validate the estimated pose against a local motion prior. 
We establish a Constant-Velocity (CV) motion prior using a short-term sliding window $\mathcal{W}$ of recent keyframes. Let $\Delta T_k = \{ \Delta R_k, \Delta \mathbf{t}_k \} \in SE(3)$ denote the relative transformation at step $k$. The motion prior for the current frame $t$ is approximated by the moving average over the local window:
\begin{equation} \label{eq:kinematic_prior}
    \Delta \tilde{\mathbf{t}}_t = \frac{1}{|\mathcal{W}|-1} \sum_{i=t-|\mathcal{W}|+1}^{t-1} \Delta \mathbf{t}_i, \quad \Delta \tilde{R}_t \approx \Delta R_{t-1}.
\end{equation}

Upon receiving the raw visual estimation $\Delta T_t = \{ \Delta R_t, \Delta \mathbf{t}_t \}$, we evaluate its physical plausibility against two critical criteria:

\paragraph{Rotational Consistency.} 
Sudden severe rotations are physically impossible for large vehicles. We compute the geodesic distance $d_R$ between the estimated rotation and the motion prior in $SO(3)$ space:
\begin{equation} \label{eq:rot_consistency}
    d_R(\Delta R_t, \Delta \tilde{R}_t) = \arccos\left(\frac{\text{Tr}(\Delta \tilde{R}_t^\top \Delta R_t) - 1}{2}\right) > \tau_R,
\end{equation}
where $\tau_R$ is the maximum allowable angular deviation. A violation typically indicates planar degeneracy (\eg, a "ghost rotation" induced by ground shadows).

\paragraph{Non-Holonomic Constraint.} 
According to Ackermann steering geometry \cite{rajamani2006vehicle}, a vehicle's primary translation must align with its forward axis $\mathbf{u}_{\text{forward}}$ (typically the Z-axis in camera coordinates). To prevent physically impossible lateral side-slips or vertical jumps, we check the cosine similarity:
\begin{equation} \label{eq:non_holonomic}
    \frac{\Delta \mathbf{t}_t \cdot \mathbf{u}_{\text{forward}}}{\|\Delta \mathbf{t}_t\|_2} < \cos(\tau_\theta).
\end{equation}
If the current geometric estimation violates either the rotational consistency in Eq. (\ref{eq:rot_consistency}) or the non-holonomic bound in Eq. (\ref{eq:non_holonomic}), the frame is flagged as a motion anomaly, triggering a 3D foundation model-based dense matching activation.

\subsubsection{Foundation Model Activation}
\label{sec:vfm_fallback}
When the motion verification detects an anomaly, it typically indicates a catastrophic failure of standard sparse feature matching, often caused by severe shadows or textureless regions. 
To recover the camera pose, we introduce a dense matching activation driven by a 3D foundation model \cite{leroy2024mast3r}. 
Since continuous inference is impractical for real-time SLAM, this mechanism is strictly on-demand. It is activated only when the initial sparse matching yields insufficient inliers or when the motion verification fails. Upon activation, we sparsify the dense correspondences using a confidence-based Non-Maximum Suppression (NMS). These refined matches subsequently replace the erroneous sparse points in the Dual-Modal Pose Estimator, successfully rescuing the trajectory without sacrificing overall efficiency.

\subsubsection{Dynamic Filtering via Sparse Optical Flow}
\label{sec:dynamic_filter}
To prevent salient moving objects from interfering with camera tracking, we introduce a lightweight pruning step without relying on external semantic segmentation modules \cite{ravi2025sam2,hu2026vggt4d}. We treat the established correspondences $\mathcal{C} = \{(\mathbf{u}_{t-1}^i, \mathbf{u}_t^i)\}$ as a sparse optical flow field.

Exploiting the tight clustering of static background flows, we identify dynamic agents as statistical outliers. 
Let $f_i = \|\mathbf{u}_t^i - \mathbf{u}_{t-1}^i\|_2$ denote the $L_2$ norm of each flow vector, and $\mathcal{F}$ denote the set of all magnitudes. 
Using robust statistics, we filter out features satisfying:
\begin{equation} \label{eq:flow_filter}
    f_i > \text{Median}(\mathcal{F}) + 3 \cdot \text{MAD}(\mathcal{F}),
\end{equation}
where $\text{MAD}$ is the Median Absolute Deviation. By statistically eliminating high-velocity dynamic objects and large-displacement mismatches, this operation significantly reduces the overall outlier ratio, enhancing the effectiveness of the subsequent geometric RANSAC.

\subsection{Map Representation and Optimization}
\label{sec:mapping}

To achieve memory-efficient and accurate mapping in large-scale outdoor environments, our backend builds the map upon the voxelized representation of Scaffold-GS \cite{lu2024scaffold-gs} and the hierarchical structure from \cite{deng2025gigaslam}, continuously optimizing the scene within a local sliding window.
% Comprehensive details regarding the hierarchical mapping and rendering formulations are provided in the supplementary material.

\subsubsection{Voxelized Gaussian Representation}
We divide the 3D scene into a grid of sparse voxels, parameterized by a set of anchor points $\mathcal{A} = \{\mathbf{p}_i \in \mathbb{R}^3\}_{i=1}^N$ centered in each voxel. Each anchor $\mathbf{p}_i$ stores a neural context feature $\mathbf{f}_i \in \mathbb{R}^{32}$, a local scaling factor $\mathbf{I}_i \in \mathbb{R}^{3}$, and $k$ positional offsets $\mathbf{O}_i \in \mathbb{R}^{k\times 3}$. During rendering, MLPs dynamically decode these attributes into $k$ explicit 3D Gaussian primitives, conditioned on the camera-to-anchor distance and viewing direction. By encoding the scene into sparse voxels and neural features, this representation effectively models view-dependent effects and improves memory efficiency.

\subsubsection{Hierarchical Anchor Management}
Outdoor driving scenes exhibit vast depth ranges. Representing distant backgrounds (\eg, sky, far buildings) with dense primitives yields diminishing returns and incurs severe computational bottlenecks. Inspired by \cite{deng2025gigaslam}, we incorporate a hierarchical Level-of-Detail (LoD) architecture to manage the sparse anchor points. 

The scene is discretized into $m$ resolution levels with corresponding voxel sizes $\{\epsilon_1, \dots, \epsilon_m\}$. When new observations are integrated, the initialized 3D position $\mathbf{x}$ is assigned to a specific level $l$ based on its distance to the camera. The position of the new anchor $\mathbf{p}_i$ is then quantized to the corresponding voxel grid: $\mathbf{p}_i = \lfloor \mathbf{x} / \epsilon_l \rfloor \cdot \epsilon_l$. During rasterization, a distance-aware mask activates fine-grained anchors for nearby regions and coarse anchors for distant areas. This hierarchical management preserves visual quality while substantially reducing memory overhead.

\subsubsection{Sliding-Window Map Optimization}
To maintain real-time efficiency, we jointly optimize anchor attributes and MLP parameters within a local sliding window $\mathcal{W}$. 
% The photometric loss $\mathcal{L}_{\text{photo}}$ measures rendering errors via $L_1$ and SSIM. To enforce structural consistency, we incorporate a geometric loss $\mathcal{L}_{\text{geo}}$, and an isotropic regularization $\mathcal{L}_{\text{iso}}$ \cite{matsuki2024monogs} to penalize highly elongated primitives. The total loss is defined as:
The total mapping loss combines a photometric loss $\mathcal{L}_{\text{photo}}$ (using $L_1$ and SSIM), a geometric loss $\mathcal{L}_{\text{geo}}$, and an isotropic regularization $\mathcal{L}_{\text{iso}}$ \cite{matsuki2024monogs} to penalize highly elongated primitives:
\begin{equation} \label{eq:mapping_loss}
    \mathcal{L}_{\text{total}} = \mathcal{L}_{\text{photo}} + \lambda_g \mathcal{L}_{\text{geo}} + \lambda_{\text{iso}} \mathcal{L}_{\text{iso}},
\end{equation}
where $\lambda_g$ and $\lambda_{\text{iso}}$ are weighting coefficients.

While these operations ensure scalability, the primary challenge for large-scale 3DGS-SLAM remains strategically managing anchor points to maximize accuracy and minimize redundancy.

%%%%%%%%%%%%%%%%%%%%%%%%%%%%%%%%%%%%%%  Lifecycle-Managed Gaussian Mapping  %%%%%%%%%%%%%%%%%%%%%%%%%%%

\subsection{Lifecycle-Managed Gaussian Mapping}
\label{sec:lifecycle}
% To build a robust geometric baseline, we precisely control the on-demand insertion of new anchors at incoming keyframes. Subsequently, we dynamically refine these anchors via a scalable chunk-based mechanism, leveraging multi-view photometric errors to accurately guide densification and pruning.
To build a robust geometric baseline, we manage the Gaussian lifecycle in two stages: probabilistic insertion of new anchors at incoming keyframes, followed by a scalable, chunk-based refinement mechanism that leverages multi-view photometric errors for densification and pruning.

\subsubsection{Probabilistic Gaussian Initialization}
To balance rendering quality and memory footprint, new Gaussian anchors are instantiated only in under-reconstructed regions with high textural or geometric complexity, rather than relying on uniform random sampling \cite{yu2025opengs-slam,cheng2025s3po-gs,deng2025gigaslam}. 

Inspired by the LoG-based sampling in \cite{meuleman2025on-the-fly}, we employ the Difference of Gaussians (DoG) for robust multi-scale texture edge detection:
\begin{equation}
    \text{DoG}(u,v) = \min(\left| (G_{\sigma_1} - G_{\sigma_2}) * I(u,v) \right|, 1),
\end{equation}
where $G_\sigma$ denotes a 2D Gaussian blur kernel, and $*$ is the convolution operation.

However, purely textural metrics often fail in textureless yet geometrically complex areas (\eg, structural corners). Thus, we introduce the normalized spatial depth gradient $\overline{\|\nabla D(u,v)\|}$ as a geometric complement. For the initial frame, the sampling probability $P(u,v)$ combines textural and geometric saliency:
\begin{equation}
    P(u,v) = \min \left( \lambda \text{DoG}(u,v) + (1-\lambda) \overline{\|\nabla D(u,v)\|}, 1 \right),
\end{equation}
where $\overline{(\cdot)}$ denotes min-max normalization to $[0, 1]$, and $\lambda$ balances the two terms. 

For subsequent keyframes, to prevent redundant insertions, we render image $\hat{I}$ and depth $\hat{D}$ to compute a rendered probability map $\hat{P}(u,v)$. This acts as a spatial penalty, defining the final spawning probability via the residual:
\begin{equation}
    P_{kf}(u,v) = \max \left( P(u,v) - \hat{P}(u,v), 0 \right).
\end{equation}
To align this continuous probability map with our discrete hierarchical voxels, we apply a $3 \times 3$ Non-Maximum Suppression (NMS) over $P_{kf}$.% to suppress local redundant candidates. 
The selected 2D pixels are unprojected into 3D space and quantized into hierarchical voxel anchors according to the LoD rules defined in Section \ref{sec:mapping}.
% The selected 2D pixels are unprojected into 3D space $\mathbf{x}$ and quantized into voxel anchors $\mathbf{p}_i = \lfloor \mathbf{x} / \epsilon_l \rfloor \cdot \epsilon_l$ according to the LoD rules. 
Finally, anchor attributes are initialized using geometry-aware heuristics from \cite{meuleman2025on-the-fly}.

\begin{table}[t]
    \centering
    % --- 字体大小控制 ---
    \fontsize{7pt}{8pt}\selectfont
    % --- 行间距与列间距控制接口 ---
    \renewcommand{\arraystretch}{1.1} % 行高倍率 (默认1.0，1.1-1.2更美观)
    \setlength{\tabcolsep}{3pt}      % 列间距 (默认6pt，数据多时可缩小如3-4pt)
    
    \caption{Camera tracking results (ATE RMSE [m]) on the KITTI dataset. \textbf{LC} denotes loop closure, ``-'' indicates failure. The best results are highlighted in \textbf{bold}, and the second-best are \underline{underlined}. Our method achieves the best overall performance, remaining highly competitive even without loop closure.}
    % \vspace{-2pt}
    \label{tab:kitti_results}
    
    \begin{tabular}{l|c|c|cccccccccc}
        \toprule
        \textbf{Methods} & \textbf{LC} & \textbf{Avg.} & \textbf{00} & \textbf{02} & \textbf{03} & \textbf{04} & \textbf{05} & \textbf{06} & \textbf{07} & \textbf{08} & \textbf{09} & \textbf{10} \\
        \midrule
        \textit{Seq. frames} & - & 2169 & 4542 & 4661 & 801 & 271 & 2761 & 1101 & 1101 & 4077 & 1591 & 1201 \\
        \textit{Seq. length (m)} & - & 2012 & 3724 & 5067 & 560 & 393 & 2205 & 1232 & 649 & 3222 & 1705 & 919 \\
        \midrule
        % ORB-SLAM2 \cite{mur2017orb} (w/o LC) & \ding{55} & 69.727 & 40.65 & 47.82 & \textbf{0.94} & 1.30 & 29.95 & 40.82 & 16.04 & 43.09 & 38.77 & 5.42 \\
        ORB-SLAM2 \cite{mur2017orb-slam2}& \ding{51} & 10.8 & 7.62 & 18.6 & 1.52 & 1.34 & \underline{4.57} & 18.4 & 1.83 & 38.8 & 8.39 & 6.80 \\
        % LDSO \cite{gao2018ldso} & \ding{51} & 22.425 & 9.32 & 31.98 & 2.85 & 1.22 & 5.10 & 13.55 & 2.96 & 129.02 & 21.64 & 17.36 \\
        % DF-VO \cite{zhan2021df-vo} & \ding{55} & 16.4 & 14.4 & 19.7 & 1.00 & 1.39 & \textbf{3.61} & 3.20 & \textbf{0.98} & 7.63 & 8.36 & 3.13 \\
        % DROID-VO \cite{teed2021droid} & \ding{55} & 54.188 & 98.43 & 108.80 & 2.58 & 0.93 & 59.27 & 64.40 & 24.20 & 64.55 & 71.80 & 16.91 \\
        DROID-SLAM \cite{teed2021droid-slam} & \ding{51} & 96.5 & 121 & 239 & 13.2 & 2.7 & 85.9 & 66.4 & 43.4 & 138 & 157 & 63.3 \\
        DPVO \cite{teed2023dpvo} & \ding{55} & 57.3 & 113 & 120 & 2.55 & \underline{1.12} & 60.8 & 52.9 & 18.2 & 113 & 77.2 & 13.1 \\
        DPV-SLAM++ \cite{lipson2024dpv-slam} & \ding{51} & 22.3 & \underline{8.30} & 39.4 & 2.50 & \textbf{0.81} & 5.74 & 11.6 & 1.58 & 110 & 76.9 & 13.6 \\
        MASt3R-SLAM \cite{murai2025mast3r-slam} & \ding{51} & 89.5 & - & - & 18.1 & 88.9 & 159 & 92.3 & - & - & - & - \\
        VGGT-SLAM \cite{maggio2025vggt-slam} & \ding{51} & 152 & 190 & 296 & 169 & 13.1 & 137 & 148 & 61.2 & 207 & 179 & 126  \\
        VGGT-Long \cite{deng2025vggt-long} & \ding{51} & 22.4 & 8.63 & 52.9 & 8.74 & 4.21 & 9.84 & 4.67 & 2.66 & 73.1 & 31.8 & 27.7 \\
        DA3-Streaming \cite{deng2025vggt-long} & \ding{51} & 21.9 & 22.9 & 68.3 & 5.15 & 1.84 & 18.4 & 5.48 & 4.89 & 24.9 & 54.4 & 13.5 \\
        % DPV-SLAM++ \cite{lipson2025deep} & \ding{51} & 25.74 & 8.30 & 39.64 & 2.50 & 0.78 & 5.74 & 11.60 & 1.52 & 110.90 & 76.70 & 13.70 \\
        % MonoGS \cite{matsuki2024monogs} & \ding{55} & - & \textit{failed} & \textit{failed} & \textit{failed} & 20.75 & \textit{failed} & & \textit{failed} & \textit{failed} & \textit{failed} & \textit{failed} \\
        Splat-SLAM \cite{sandstrom2025splat-slam} & \ding{51} & 45.4 & 83.1 & - & 3.40 & 1.72 & 33.0 & 130 & 14.3 & 52.0 & 27.4 & 63.5 \\
        OnTheFly-NVS \cite{meuleman2025on-the-fly} & \ding{55} & 77.5 & 116 & 169 & 18.1 & 9.08 & 78.4 & 82.2 & 39.3 & - & 114 & 46.4 \\
        % OpenGS-SLAM \cite{yu2025opengs-slam} & \ding{51} & - &  &  &  &  &  &  &  &  & & \\
        S3PO-GS \cite{cheng2025s3po-gs} & \ding{55} & 77.1 & 94.1 & 264 & 22.9 & 14.5 & 56.5 & 49.4 & 20.9 & 60.9 & 91.9 & 96.8 \\
        Vings-Mono \cite{wu2025vings-mono} & \ding{51} & 74.3 & 100 & 206 & 9.70 & 1.67 & 73.0 & 93.8 & 31.9 & 124 & 76.7 & 25.2 \\
        GigaSLAM \cite{deng2025gigaslam} & \ding{51} & 5.46 & 11.2 & \underline{11.3} & \underline{1.41} & 1.58 & 4.92 & \underline{2.11} & 3.17 & \underline{13.5} & \underline{3.24} & \textbf{2.15} \\
        \midrule
        \textbf{Ours (w/o LC)} & \ding{55} & \underline{4.26} & 8.94 & \underline{11.3} & 1.18 & 1.53 & 4.86 & 2.66 & \underline{1.80} & \textbf{4.90} & \textbf{3.06} & \underline{2.38} \\
        \textbf{Ours (w/ LC)} & \ding{51} & \textbf{3.46} & \textbf{6.23} & \textbf{8.70} & \textbf{1.18} & 1.53 & \textbf{4.32} & \textbf{1.26} & \textbf{1.06} & \textbf{4.90} & \textbf{3.06} & \underline{2.38} \\
        \bottomrule
    \end{tabular}
    % \vspace{-5pt}
\end{table}

\subsubsection{Multi-View Consistency Management}
Standard 3DGS relies on positional gradients for densification, which introduces redundancy and is highly sensitive to pose noise in outdoor SLAM. Recent methods \cite{wu2025vings-mono} introduce local heuristics for Gaussian management, but they ignore multi-view consistency, frequently leading to erroneous pruning. 
% To ensure structural consistency, we utilize multi-view reprojection errors \cite{ren2025fastgs} to accurately guide anchor densification and pruning.
Inspired by \cite{ren2025fastgs}, we utilize multi-view reprojection errors to accurately guide anchor densification and pruning.
% Standard 3DGS relies on view-space positional gradients for densification, which are highly sensitive to pose noise and often introduce redundancy in outdoor SLAM. Recent methods \cite{qin2018vins-mono} introduce local heuristics for Gaussian management, but they ignore multi-view consistency, frequently leading to erroneous pruning. To ensure structural consistency, we leverage multi-view reprojection errors \cite{ren2025fastgs} to guide anchor densification and pruning.

\begin{table}[t]
    \centering
    % --- 字体大小控制 ---
    \fontsize{7pt}{8pt}\selectfont
    
    % --- 行间距与列间距控制 ---
    \renewcommand{\arraystretch}{1.1} 
    \setlength{\tabcolsep}{2.4pt}
    
    \caption{Camera Tracking Results (ATE RMSE [m]) on the KITTI-360 Dataset. Our method consistently ranks within the top two across all sequences, securing the best performance in the majority of them.} %\textbf{LC} denotes loop closure.}
    % \vspace{-2pt}
    \label{tab:kitti360_results}
    
    \begin{tabular}{l|c|c|ccccccccc}
        \toprule
        \textbf{Methods} & \textbf{LC} & \textbf{Avg.} & \textbf{0000} & \textbf{0002} & \textbf{0003} & \textbf{0004} & \textbf{0005} & \textbf{0006} & \textbf{0007} & \textbf{0009} & \textbf{0010} \\
        \midrule
        % 已填入数据 (四舍五入取整)
        \textit{Seq. frames} & - & 7898 & 10505 & 13888 & 1010 & 11047 & 6289 & 9185 & 2890 & 13246 & 3022 \\
        \textit{Seq. length (m)} & - & 6984 & 8403 & 11616 & 1379 & 9975 & 4691 & 7980 & 4888 & 10580 & 3343 \\
        \midrule
        ORB-SLAM2 \cite{mur2017orb-slam2} & \ding{51} & 47.1 & 45.5 & 32.3 & 11.5 & 72.1 & 24.8 & 43.9 & 134 & - & \textbf{12.6} \\
        % DF-VO \cite{zhan2021df-vo} & \ding{55} & & & & & & & & & & \\
        DROID-SLAM \cite{teed2021droid-slam} & \ding{51} & 156 & 110 & 233 & 10.4 & 169 & 139 & 113 & 433 & 165 & 31.9 \\
        % DPVO \cite{teed2023dpvo} & \ding{55} & 47.5 & 50.7 & 57.3 & 2.97 & 50.3 & 68.1 & 39.2 & 35.3 & 90.5 & 32.9 \\
        DPV-SLAM++ \cite{lipson2024dpv-slam} & \ding{51} & \underline{32.5} & \underline{27.8} & 30.6 & \textbf{3.01} & \underline{40.5} & 20.4 & 39.6 & \textbf{36.5} & \underline{65.2} & \underline{30.7}  \\
        % MASt3R-SLAM \cite{murai2025mast3r-slam} & \ding{51} & & & & & & & & & & \\
        % VGGT-SLAM \cite{maggio2025vggt-slam} & \ding{51} & & & & & & & & & & \\
        VGGT-Long \cite{deng2025vggt-long} & \ding{51} & 344 & 305 & 466 & 17.6 & 703 & 138 & 248 & 491 & 241 & 487 \\
        DA3-Streaming \cite{deng2025vggt-long} & \ding{51} & 243 & 233 & 480 & 21.9 & 318 & 192 & 256 & 371 & 177 & 140 \\
        % MonoGS \cite{matsuki2024monogs} & \ding{55} & & & & & & & & & & \\
        % Splat-SLAM \cite{sandstrom2025splat-slam} & \ding{51} & & & & & & & & & & \\
        OnTheFly-NVS \cite{meuleman2025on-the-fly} & \ding{55} & 165 & 164 & 247 & 32.3 & 179 & 132 & 182 & 257 & - & 127 \\
        Vings-Mono \cite{wu2025vings-mono} & \ding{51} & 171 & 168 & - & 77.6 & - & - & - & 298 & - & 142 \\
        % OpenGS-SLAM \cite{yu2025opengs-slam} & \ding{51} & & & & & & & & & & \\
        S3PO-GS \cite{cheng2025s3po-gs} & \ding{55} & 167 & - & - & 55.5 & - & 83.5 & - & 260 & - & 271 \\
        GigaSLAM \cite{deng2025gigaslam} & \ding{51} & 64.7 & 42.6 & 35.1 & 24.4 & 206 & \underline{15.0} & \underline{37.6} & 59.5 & 106 & 56.2 \\
        \midrule
        % \textbf{Ours (w/o LC)} & \ding{55} & & & & & & & & & & \\
        \textbf{Ours} & \ding{51} & \textbf{25.7} & \textbf{23.9} & \textbf{17.7} & \underline{17.3} & \textbf{36.7} & \textbf{9.54} & \textbf{33.7} & \underline{50.0} & \textbf{23.2}  & \underline{19.5} \\
        \bottomrule
    \end{tabular}
    % \vspace{-5pt}
\end{table}

Since global multi-view evaluation is computationally prohibitive for large-scale SLAM, we divide the historical keyframe sequence into local chunks of size $C$.  Within each chunk $c$, we compute a densification score $S_{d}^{(i, c)}$ (identifying under-reconstructed regions) and a pruning score $S_{p}^{(i, c)}$ (measuring error contribution) for each anchor $i$. Detailed formulations are provided in the supplementary material. To robustly update the global voxel grid, we aggregate these chunk-level scores using two distinct strategies:

\paragraph{Global Densification Rules.} 
%To ensure no local structural defect is overlooked, 
Anchor $i$ is globally marked for densification if its score exceeds a threshold $\tau_{d}$ in \emph{any} individual chunk:
\begin{equation}
    S_{d}(i) = \bigvee_{c} \left( S_{d}^{(i, c)} > \tau_{d} \right).
\end{equation}

\paragraph{Global Pruning Rules.} Since $S_{p}$ reflects rendering degradation, we evaluate the worst-case impact of anchor $i$ by taking its maximum score across all chunks:
\begin{equation}
    S_{p}(i) = \max_{c} \left( S_{p}^{(i, c)} \right).
\end{equation}
Anchor $i$ is pruned if $S_{p}(i) > \tau_p$, eliminating Gaussian floaters while preserving valid structures across complex views.

%%======================================4-Experiments============================================
\section{Experiments}
\label{sec:experiments}

\subsection{Experimental Setup}
\label{sec:exp_setup}

% \subsubsection{Datasets and Metrics}
\noindent \textbf{Datasets and Metrics.}
% We evaluate our system on three outdoor datasets: Waymo, KITTI, and KITTI-360. 
% Waymo contains diverse street-view sequences of nearly 200 frames; we follow the standard protocol evaluating on nine selected scenes. 
% KITTI provides longer sequences up to 4,661 frames ($5$\,km), while KITTI-360 features ultra-long trajectories extending up to 13,888 frames ($11.6$\,km), posing extreme challenges for monocular SLAM. 
% Note that we omit the KITTI-01 sequence, as its extreme highway speeds and lack of trackable near-field features cause universal degeneracy across most monocular baselines.
% For tracking accuracy, we report the Absolute Trajectory Error (ATE) RMSE [m] on keyframes. 
% Mapping quality is assessed using standard novel view synthesis metrics: PSNR, SSIM, and LPIPS. 
% Furthermore, to evaluate system efficiency for 3DGS-based methods, we report training time, peak VRAM usage, the total number of Gaussians, and final storage size.
We evaluate our system on three outdoor datasets: Waymo \cite{sun2020waymo} (nine 200-frame sequences following \cite{cheng2025s3po-gs}), KITTI \cite{geiger2013kitti} (up to 4,661 frames, 5 km), and the highly challenging KITTI-360 \cite{liao2022kitti-360} (up to 13,888 frames, 11.6 km). We omit KITTI-01, as its extreme highway speeds and lack of trackable near-field features cause universal monocular degeneracy.

For tracking accuracy, we report the Absolute Trajectory Error (ATE) RMSE [m] on keyframes.
Mapping quality is assessed via PSNR, SSIM, and LPIPS.
To evaluate system efficiency for 3DGS-based methods, we report training time, peak VRAM usage, number of Gaussians, and map storage size.

% \subsubsection{Baseline Methods}
\begin{table}[t]
    \centering
    % --- 字体大小控制 ---
    \fontsize{6pt}{7.5pt}\selectfont % 第一个字体大小；第二个控制文字与文字之间的紧密程度（特别是当一个格子里有两行字时）
    
    % --- 行间距与列间距控制 ---
    \renewcommand{\arraystretch}{1.1} % 控制格子与格子之间的宽松程度（撑大单元格的上下留白）
    \setlength{\tabcolsep}{2pt} % 删了一列后，列间距可以稍微宽一点点(2pt -> 2.5pt)
    
    \caption{Camera Tracking Results (ATE RMSE [m]) on the Waymo Dataset.}
    % \vspace{-2pt}
    \label{tab:waymo_results}
    
    % 移除了 LC 列，现在是 1(Method) + 1(Avg) + 9(Scenes) = 11列
    \begin{tabular}{l|c|ccccccccc}
        \toprule
        \textbf{Methods} & \textbf{Avg.} & \textbf{13476} & \textbf{100613} & \textbf{106762} & \textbf{132384} & \textbf{152706} & \textbf{153495} & \textbf{158686} & \textbf{163453} & \textbf{405841} \\
        \midrule
        % \textit{Seq. frames} & 198 & 197 & 198 & 198 & 199 & 199 & 199 & 198 & 198 & 199 \\
        % \textit{Seq. length (m)} & & & & & & & & & & \\
        % \midrule
        ORB-SLAM2 \cite{mur2017orb-slam2} & 3.95 & 1.08 & 4.77 & 8.01 & 4.55 & 6.79 & 2.21 & 4.25 & 2.59 & 1.31 \\
        % DF-VO \cite{zhan2021df-vo} & & & & & & & & & & \\
        DROID-SLAM \cite{teed2021droid-slam} & 10.4 & 6.71 & 3.95 & 3.45 & 19.4 & 5.18 & 6.42 & 11.5 & 33.1 & 3.92 \\
        DPVO \cite{teed2023dpvo} & 0.85 & 0.93 & 0.59 & \textbf{0.29} & 0.68 & 0.72 & 1.30 & 1.12 & 0.99 & 1.04 \\
        % DPV-SLAM \cite{lipson2024dpv-slam} & & & & & & & & & & \\
        MASt3R-SLAM \cite{murai2025mast3r-slam} & 1.87 & 3.15 & 0.86 & 0.98 & 1.03 & 0.82 & 4.30 & 1.87 & 2.46 & 1.35 \\
        VGGT-SLAM \cite{maggio2025vggt-slam} & 5.81 & 5.54 & 6.32 & 3.55 & 4.64 & 5.46 & 6.93 & 10.0 & 5.58 & 4.27 \\
        VGGT-Long \cite{deng2025vggt-long} & 1.46 & 0.71 & 1.31 & 2.64 & 1.67 & 0.62 & 1.17 & 1.57 & 1.92 & 1.61 \\
        DA3-Streaming \cite{deng2025vggt-long} & 1.35 & \underline{0.61} & 1.08 & 0.77 & \underline{0.42} & 1.76 & 2.21 & 1.96 & 1.70 & 0.92 \\
        % MonoGS \cite{matsuki2024monogs} & & & & & & & & & & \\
        Splat-SLAM \cite{sandstrom2025splat-slam} & \underline{0.59} & 0.62 & 0.40 & 0.41 & \textbf{0.24} & 0.63 & 1.20 & 0.66 & \underline{0.65} & 0.54 \\
        OnTheFly-NVS \cite{meuleman2025on-the-fly} & 2.66 & 5.65 & 1.60 & 1.17 & 1.43 & 2.24 & 3.05 & 1.74 & 4.68 & 2.36 \\
        OpenGS-SLAM \cite{yu2025opengs-slam} & 1.38 & 2.74 & \textbf{0.33} & 0.89 & 0.78  & 0.62 & 1.62 & 1.18 & 3.31 & 0.93 \\
        S3PO-GS \cite{cheng2025s3po-gs} & 0.92 & 1.04 & 0.76 & 0.37 & 0.62 & 0.79 & 2.11 & 0.69 & 1.42 & \underline{0.51} \\
        Vings-Mono \cite{wu2025vings-mono} & 2.28 & 2.69 & 1.11 & 3.89 & 3.88 & 1.69 & 2.09 & 2.15 & 2.26 & 0.73 \\
        GigaSLAM \cite{deng2025gigaslam} & 0.87 & 2.54 & \underline{0.34} & 0.38 & 0.97 & \underline{0.48} & \textbf{0.64} & \underline{0.63} & 1.32 & 0.52 \\
        \midrule
        % 只保留 Ours 最终结果
        \textbf{Ours} & \textbf{0.51} & \textbf{0.58} & \textbf{0.33} & \underline{0.36} & 0.55 & \textbf{0.46} & \underline{0.68} & \textbf{0.59} & \textbf{0.59} & \textbf{0.47} \\
        \bottomrule
    \end{tabular}
    % \vspace{-5pt}
\end{table}

\begin{figure*}[t]
% \centerline{\includegraphics[width=1.0\textwidth, trim=0cm 8cm 11cm 0cm, clip]{figs/排版占位.pdf}}
\centerline{\includegraphics[width=0.95\textwidth]{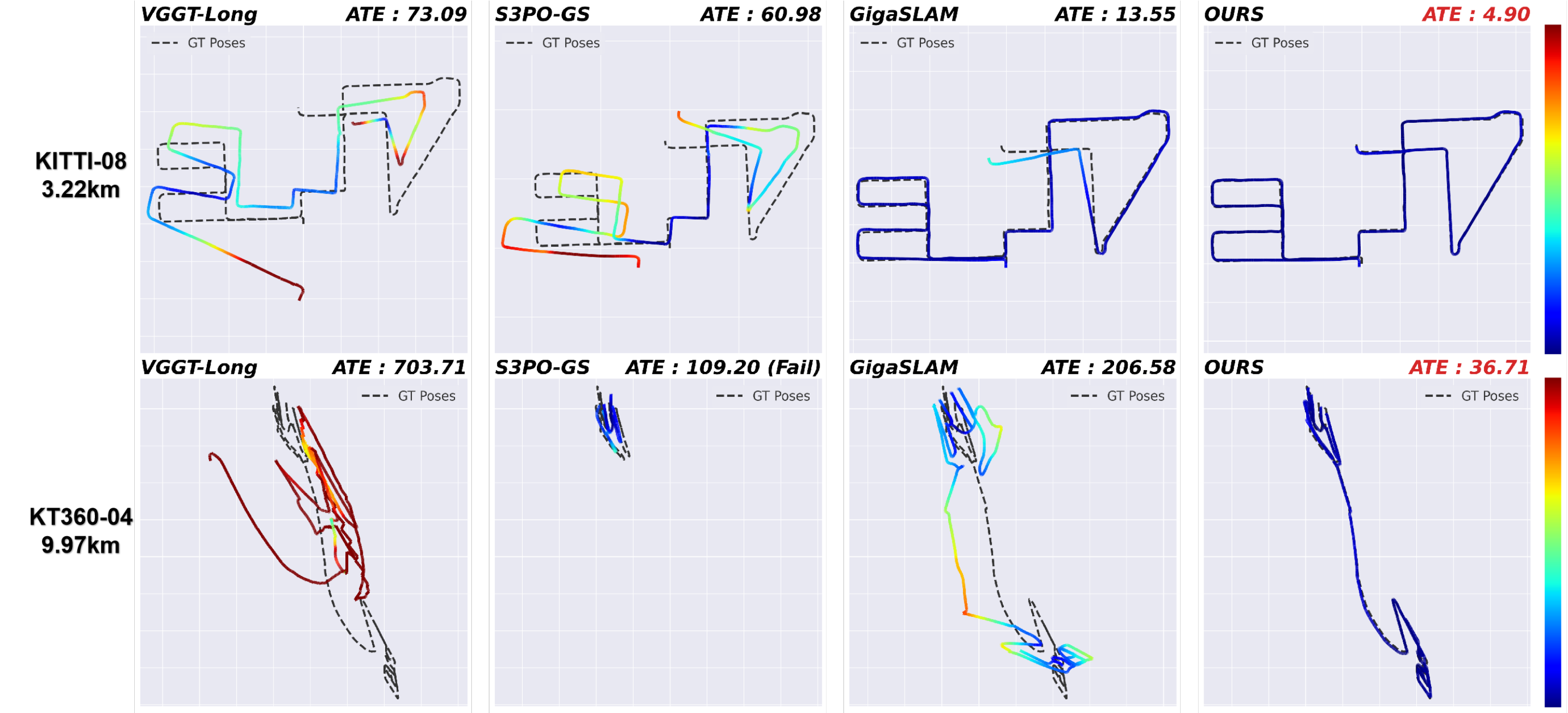}}
% \vspace{-2pt}
\caption{\textbf{Tracking trajectories.} Our method maintains robust pose estimation and global consistency across long-term sequences, whereas other baselines inevitably suffer from significant drift in localized challenging segments.}
% \vspace{-5pt}
\label{trajectory}
\end{figure*}

\noindent \textbf{Baseline Methods.} 
We compare our approach with two distinct categories of state-of-the-art baselines. 
The first category is long-sequence VO/SLAM systems: ORB-SLAM2 \cite{mur2017orb-slam2}, DF-VO \cite{zhan2021df-vo}, DROID-SLAM \cite{teed2021droid-slam}, DPVO \cite{teed2023dpvo}, DPV-SLAM \cite{lipson2024dpv-slam}, MASt3R-SLAM \cite{leroy2024mast3r}, VGGT-SLAM \cite{maggio2025vggt-slam}, VGGT-Long \cite{deng2025vggt-long} and DA3-Streaming \cite{deng2025vggt-long}. 
The second category is monocular 3DGS-SLAM frameworks: Splat-SLAM \cite{sandstrom2025splat-slam}, OnTheFly-NVS \cite{meuleman2025on-the-fly}, Vings-Mono \cite{wu2025vings-mono}, OpenGS-SLAM \cite{yu2025opengs-slam}, S3PO-GS \cite{cheng2025s3po-gs}, and GigaSLAM \cite{deng2025gigaslam}. 
Particularly for GigaSLAM, which relies on an extremely time-consuming color refinement process, we restrict its iterations to an acceptable bound to ensure fair efficiency comparisons.

% \subsubsection{Implementation Details}
\noindent \textbf{Implementation Details.} 
Our system is implemented in PyTorch and evaluated on single NVIDIA RTX A6000 GPU. 
We employ MASt3R \cite{leroy2024mast3r} as the dense matching model for frontend fallback mechanism, and a metric depth model \cite{piccinelli2025unidepthv2} as geometric prior. 
% To ensure global consistency, we integrate the classic DBoW2 module for loop closure detection, followed by Sim(3) pose graph optimization. 
% Crucially, upon loop closure, the corresponding rigid transformations are explicitly applied to align the 3D Gaussian map. 
For global consistency, we integrate DBoW2 \cite{galvez2012bags} loop closure and Sim(3) pose graph optimization, followed by rigid transformations to align the 3D Gaussian map.
Further details on loop closure, keyframe selection, and hyperparameters are provided in the supplementary material.

\begin{table*}[t]
    \centering
    % --- 全局控制接口 ---
    \fontsize{7pt}{8pt}\selectfont % 字号 / 行距
    \renewcommand{\arraystretch}{1.1}   % 行高倍率
    \setlength{\tabcolsep}{2pt}       % 列间距 (跨栏表格空间较足，可以设大一点)

    \caption{\textbf{Rendering Results on three Datasets.} Only 3DGS-based methods are compared. We report the average results across all evaluated scenes. Our method achieves the highest rendering quality across all three datasets.}
    \label{tab:rendering_results}
    
    \begin{tabular}{l|ccc|ccc|ccc}
        \toprule
        \multirow{2}{*}{\textbf{Method}} & \multicolumn{3}{c|}{\textbf{Waymo}} & \multicolumn{3}{c|}{\textbf{KITTI}} & \multicolumn{3}{c}{\textbf{KITTI-360}} \\ 
        \cmidrule(lr){2-4} \cmidrule(lr){5-7} \cmidrule(lr){8-10}
        & PSNR$\uparrow$ & SSIM$\uparrow$ & LPIPS$\downarrow$ & PSNR$\uparrow$ & SSIM$\uparrow$ & LPIPS$\downarrow$ & PSNR$\uparrow$ & SSIM$\uparrow$ & LPIPS$\downarrow$ \\ 
        \midrule
        % --- Baseline: 3DGS-based Methods ---
        % MonoGS \cite{matsuki2024monogs} & & & & & & & & & \\
        Splat-SLAM \cite{sandstrom2025splat-slam} & 25.34 & 0.832 & 0.378 & 17.22 & 0.611 & 0.620 & - & - & - \\
        OnTheFly-NVS \cite{meuleman2025on-the-fly} & 25.16 & 0.830 & \underline{0.349} & 16.20 & 0.541 & 0.520 & 14.19 & 0.493 & 0.577 \\
        Vings-Mono \cite{wu2025vings-mono} & 19.25 & 0.623 & 0.587 & 15.07 & 0.556 & 0.544 & 14.95 & 0.502 & 0.711 \\ 
        OpenGS-SLAM \cite{yu2025opengs-slam} & 24.83 & 0.816 & 0.415 & - & - & - & - & - & - \\
        S3PO-GS \cite{cheng2025s3po-gs} & \underline{26.84} & \underline{0.844} & 0.367 & 15.40 & 0.494 & 0.686 & 13.76 & 0.468 & 0.730 \\
        GigaSLAM \cite{deng2025gigaslam} & 26.12 & 0.822 & 0.472 & \underline{20.99} & \underline{0.650} & \underline{0.504} & \underline{20.92} & \underline{0.661} & \underline{0.516} \\
        \midrule
        % --- Ours ---
        \textbf{Ours} & \textbf{28.36} & \textbf{0.862} & \textbf{0.339} & \textbf{22.52} & \textbf{0.693} & \textbf{0.437} & \textbf{21.60} & \textbf{0.671} & \textbf{0.476} \\
        \bottomrule
    \end{tabular}
    % \vspace{-5pt} % 如果想减少表格和正文的距离，可以取消注释
\end{table*}

\begin{table*}[t]
    \centering
    \fontsize{6.5pt}{7.5pt}\selectfont 
    \renewcommand{\arraystretch}{1.1} 
    \setlength{\tabcolsep}{2.4pt} 
    % \belowrulesep=0pt\aboverulesep=0pt
    \caption{\textbf{Efficiency Analysis.} Results on Waymo are averaged across sequences ($\approx$200 frames), while KITTI \textit{Seq. 00} and KITTI-360 \textit{Seq. 0005} are selected to test long-term scalability. \textbf{T}: Runtime (min); \textbf{Mem}: Peak VRAM (GB); \textbf{Size}: Map Storage (MB); \textbf{\#GS}: Number of Gaussians. \textit{OOM} denotes Out-Of-Memory.}
    \label{tab:efficiency}
    
    \begin{tabular}{l|cccc|cccc|cccc}
        \toprule
        \multirow{2}{*}{\textbf{Method}} & \multicolumn{4}{c|}{\textbf{Waymo (Avg.)}} & \multicolumn{4}{c|}{\textbf{KITTI (Seq. 00)}} & \multicolumn{4}{c}{\textbf{KITTI-360 (Seq. 0005)}} \\ 
        \cmidrule(lr){2-5} \cmidrule(lr){6-9} \cmidrule(lr){10-13}
        & T & Mem & \#GS & Size & T & Mem & \#GS & Size & T & Mem & \#GS & Size \\ 
        \midrule
        % MonoGS & & & & & & & & & & & & \\
        Splat-SLAM & 40.3 & 16.4 & 420K & 96.2 & OOM & OOM & OOM & OOM & OOM & OOM & OOM & OOM \\
        OnTheFly-NVS & \textbf{1.87} & \underline{8.3} & 528K & 124 & \textbf{17.1} & \textbf{8.3} & 16.6M & 3925 & \textbf{22.4} & \textbf{11.2} & 31.1M & 7336 \\
        Vings-Mono & 9.77 & \textbf{3.5} & 1.1M & 262 & 229 & 21.5 & 4.18M & 1018 & Fail & Fail & Fail & Fail \\
        OpenGS-SLAM & 33.6 & 15.2 & \underline{306K} & 48.6 & OOM & OOM & OOM & OOM & OOM & OOM & OOM & OOM \\
        S3PO-GS & 31.1 & 12.8 & 320K & 50.1 & 320 & 25.0 & \underline{2.89M} & \underline{582} & 462 & 38.5 & \underline{3.21M} & \underline{560} \\
        GigaSLAM & 4.67 & 21.8 & 365K & 75.6 & 125 & 36.9 & 4.01M & 1207 & 229 & 47.2 & 6.09M & 2009 \\
        \midrule
        \textbf{Ours} & \underline{4.52} & 8.6 & \textbf{180K} & \textbf{37.3} & \underline{70.0} & \underline{18.1} & \textbf{1.12M} & \textbf{325} & \underline{127} & \underline{23.7} & \textbf{1.14M} & \textbf{380} \\
        \bottomrule
    \end{tabular}
    % \vspace{-5pt}
\end{table*}

\subsection{Results and Comparisons}
\label{sec:results}

\noindent \textbf{Camera Tracking.} We evaluate the tracking robustness of our system across varying sequence lengths. On the short-sequence Waymo dataset (Table \ref{tab:waymo_results}), our method achieves the lowest average ATE (0.51m) and consistently ranks in the top two across most scenes. For the medium-scale KITTI dataset (Table \ref{tab:kitti_results}), our approach maintains the lowest trajectory error (Avg. 3.46m). Notably, even without global loop closure (\textit{w/o LC}), our visual odometry remains highly competitive against full SLAM systems, validating the effectiveness of our motion-adaptive hybrid tracking module. 

Furthermore, on the ultra-long, highly challenging KITTI-360 dataset (Table \ref{tab:kitti360_results}), our system demonstrates unparalleled robustness. It secures the best average ATE (25.7m) and ranks within the top two across all individual sequences, successfully avoiding the catastrophic drifts that frequently plague other baselines. The qualitative trajectory visualizations (Fig. \ref{trajectory}) further corroborate our global consistency over multi-kilometer runs.

\begin{figure*}[t]
% \centerline{\includegraphics[width=1.0\textwidth, trim=0cm 8cm 11cm 0cm, clip]{figs/排版占位.pdf}}
\centerline{\includegraphics[width=1.0\textwidth]{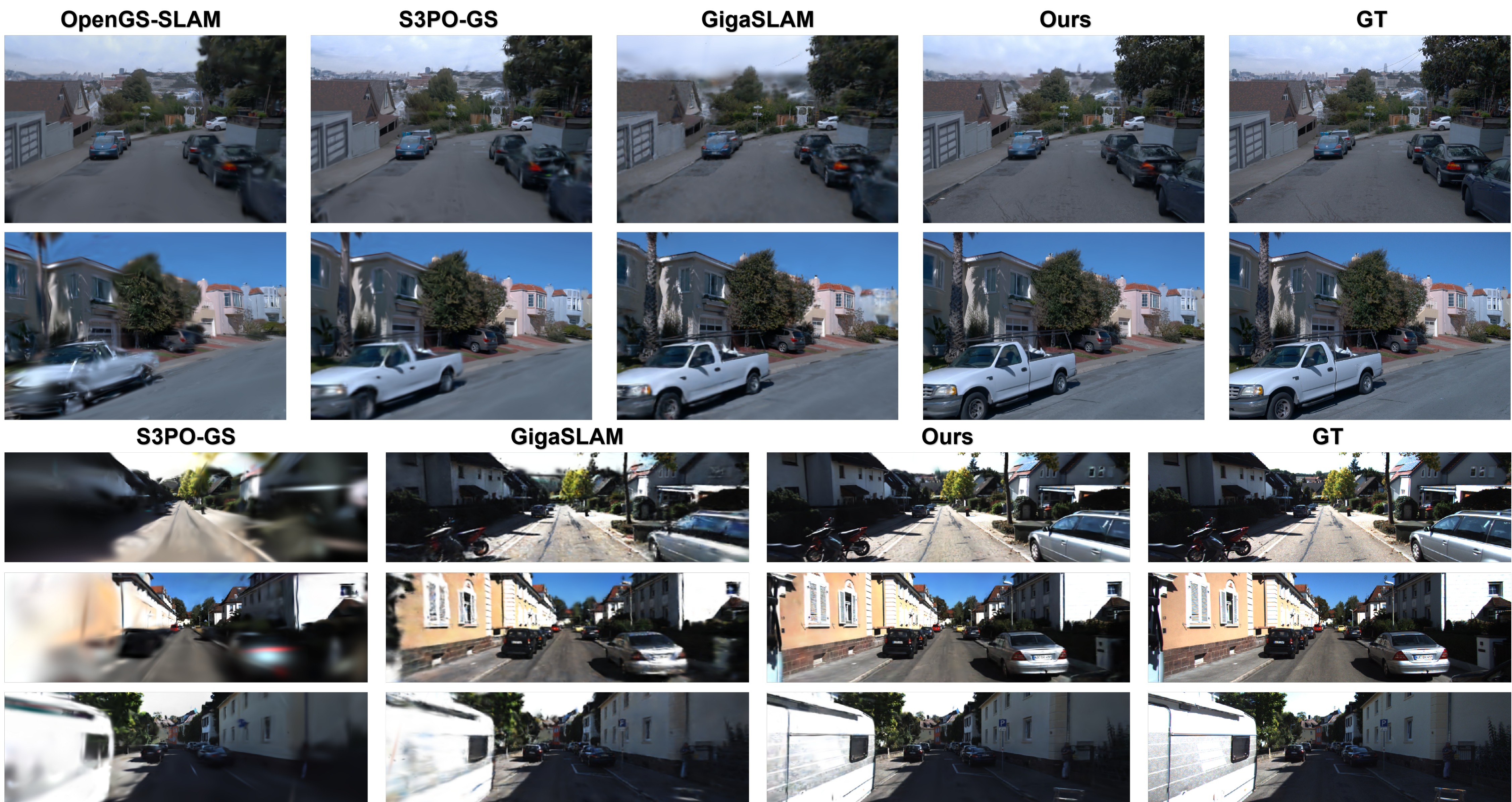}}
% \vspace{-3pt}
\caption{\textbf{Qualitative rendering comparisons on Waymo (top 2) and KITTI (bottom 3).} Our method preserves superior structural and textural details in both near-field and distant regions, whereas other baselines struggle with noticeable blurring and artifacts.}
% \vspace{-5pt}
\label{render}
\end{figure*}

\noindent \textbf{Mapping and Efficiency.} Table \ref{tab:rendering_results} demonstrates that our method yields the highest rendering quality across all three datasets (\eg, 28.36 dB PSNR on Waymo). Qualitative results (Fig. \ref{render}) further verify our superior preservation of high-frequency details in both near-field textures and distant structures. 

Table \ref{tab:efficiency} reveals the underlying efficiency behind this performance. Systems like OnTheFly-NVS and Vings-Mono rely on CPU-GPU scheduling to manage memory pressure. While OnTheFly-NVS achieves extremely fast speeds, both methods suffer from massive Gaussian redundancy (\eg, 31.1M primitives for OnTheFly-NVS on KITTI-360) and significantly degraded rendering quality. In contrast, our lifecycle-managed Gaussian mapping fundamentally compresses the scene representation. We maintain the strictly lowest primitive count (only 1.14M on KITTI-360) and minimal storage (380 MB), alongside highly competitive VRAM consumption (23.7 GB) and an acceptable runtime. This optimal trade-off between fidelity and resource constraints is crucial for seamlessly scaling to kilometer-scale outdoor environments.

\subsection{Ablation Study}
\label{sec:ablation}

% Table 1: Frontend (Transposed)
\begin{table*}[t]
\centering
\fontsize{7pt}{8pt}\selectfont
\renewcommand{\arraystretch}{1.1}
\setlength{\tabcolsep}{2.5pt}
\caption{Ablation study of Motion-Adaptive Hybrid Tracking module on KITTI-07.}
% \vspace{-5pt}
\begin{tabular}{l|ccccc}
\toprule
Metric & Only $\mathcal{M}_E$ & Only $\mathcal{M}_P$ & w/o Dyn. Filter & w/o Activation & \textbf{Ours (Full)} \\
\midrule
ATE RMSE (m) $\downarrow$ & 2.15 & 3.64 & 1.81 & 2.32 & \textbf{1.06} \\
PSNR (dB) $\uparrow$     & 21.9 & 21.2 & 22.3 & 22.1 & \textbf{22.7} \\
\bottomrule
\end{tabular}
\label{ablation1}
% \vspace{-5pt}
\end{table*}

% Table 2: Initialization
\begin{table*}[t]
\centering
\fontsize{7pt}{8pt}\selectfont
\renewcommand{\arraystretch}{1.1}
\setlength{\tabcolsep}{3pt}
\caption{Ablation of Gaussian initialization strategies on KITTI-02.}
% \vspace{-5pt}
\begin{tabular}{l|cccc}
\toprule
Strategy & PSNR $\uparrow$ & \#GS (Millions)& VRAM (GB) & Time (min) \\
\midrule
Random Sampling      & 21.84 & 365 & 44 & 120 \\
Ours (Canny instead of DoG)  & 21.05 & 128 & 18 & 85 \\
Ours (LoG instead of DoG)   & 21.62 & 142 & 20 & 90 \\
Ours w/o Depth Grad. & 21.56 & 131 & 19 & 86 \\
Ours w/o NMS         & 21.79 & 202 & 24 & 100 \\
\rowcolor{gray!10} \textbf{Ours (DoG + Depth Grad.)} & \textbf{21.77} & \textbf{135} & \textbf{19} & \textbf{87} \\
\bottomrule
\end{tabular}
% \vspace{-2pt}
\label{ablation2}
\end{table*}

% Table 3: Lifecycle Management
\begin{table*}[t]
\centering
\fontsize{7pt}{8pt}\selectfont
\renewcommand{\arraystretch}{1.1}
\setlength{\tabcolsep}{4pt}
\caption{Ablation of multi-view consistent Management on KITTI-02.}
% \vspace{-5pt}
\begin{tabular}{ccc|ccc}
\toprule
Densify & Prune & Chunk & PSNR $\uparrow$ & \#GS (Millions) & VRAM (GB) \\
\midrule
$\times$   & $\times$   & -          & 21.27 & 149 & 20  \\ 
$\times$   & \checkmark & \checkmark & 21.16 & 108 & 17  \\ 
\checkmark & $\times$   & \checkmark & 21.71 & 166 & 21 \\ 
\checkmark & \checkmark & $\times$   & \textbf{OOM}     & \textbf{OOM}  & \textbf{OOM}  \\ 
\rowcolor{gray!10} \checkmark & \checkmark & \checkmark & \textbf{21.77} & \textbf{135} & \textbf{19}  \\ 
\bottomrule
\end{tabular}
% \vspace{-3pt}
\label{ablation3}
\end{table*}

To validate our system design, we conduct ablation studies on the KITTI dataset, evaluating both tracking and mapping metrics.

\noindent \textbf{Motion-Adaptive Hybrid Tracking.} As shown in Table \ref{ablation1}, relying on a single geometric solver ($\mathcal{M}_E$ or $\mathcal{M}_P$) yields significant trajectory drift due to environmental degeneracies. Disabling dynamic filtering exacerbates errors induced by moving objects and severe mismatches. Crucially, removing the foundation model activation leaves the system vulnerable to extreme visual degradation, substantially increasing the ATE. This resilient tracking directly establishes a robust geometric foundation for the mapping, as the full hybrid tracking module directly translates to the highest rendering fidelity (22.7 dB).

\noindent \textbf{Gaussian Initialization.} Table \ref{ablation2} evaluates our initialization strategy against alternative operators. Canny, as a single-scale edge detector, inserts insufficient primitives, degrading the PSNR. Compared to the LoG operator, which is highly sensitive to outdoor high-frequency noise, DoG extracts multi-scale textural features more robustly. While uniform random sampling or omitting Non-Maximum Suppression (w/o NMS) yields marginal PSNR gains, they cause prohibitive explosions in primitive count, VRAM (\eg, 44 GB), and runtime. Our combined strategy (DoG + Depth Grad.) demonstrates its superiority by achieving the optimal accuracy-efficiency trade-off.

\noindent \textbf{Lifecycle-Managed Mapping.} Table \ref{ablation3} demonstrates the necessity of our multi-view consistent management. Attempting to compute multi-view scores globally without historical chunking (\textit{w/o Chunk}) immediately exhausts GPU memory (OOM) on long sequences. Densification effectively recovers under-reconstructed regions omitted during initialization, significantly boosting the PSNR. Conversely, our pruning mechanism reliably identifies and eliminates redundant floaters. This not only compresses the map substantially (from 166M to 135M Gaussians) but also refines the overall rendering quality, proving the effectiveness of our continuous optimization pipeline.

% \paragraph{Frontend Robustness.}
% We isolate the contributions of our Kinematics-Aware Adaptive Frontend. 
% (1) \textbf{w/o Dual-Modal Solver}: Relying solely on the Essential matrix or PnP solver leads to a $\text{XX}\%$ increase in ATE, confirming the necessity of GRIC-based dynamic routing in planar scenes. 
% (2) \textbf{w/o Kinematic Fallback}: Disabling the sliding-window check and the MASt3R fallback causes catastrophic tracking failures (divergence) in scenes with severe shadows or pure rotations. 
% (3) \textbf{w/o Dynamic Filtering}: Without flow-based pruning, moving vehicles corrupt the static scene assumption, noticeably degrading local pose precision.

% \paragraph{Gaussian Lifecycle Management.}
% We further evaluate the backend mapping components. 
% (1) \textbf{w/o Probabilistic Init}: Replacing our depth-gradient and DoG-aware initialization with uniform random sampling drops the PSNR by $\text{XX}$ dB and introduces noticeable holes in textureless regions. 
% (2) \textbf{w/o Multi-View Pruning}: Disabling the chunk-based degradation score evaluation results in the accumulation of massive redundant floaters, bloating the VRAM usage by $\text{XX}\times$ without any gain in rendering quality.

%%======================================5-Conclusion============================================
\section{Conclusion}
In this paper, we presented a novel monocular 3DGS-SLAM framework specifically engineered to tackle the complexities of kilometer-scale outdoor environments. By identifying the critical bottlenecks of pure visual tracking and volumetric rendering in unbounded scenes, we developed a system that fundamentally prioritizes both resilience and efficiency. Our motion-adaptive hybrid tracking module effectively mitigates local tracking collapses by intelligently activating dense foundation models only when geometric solvers fail. Concurrently, the proposed lifecycle-managed Gaussian mapping strategy strictly regulates the growth of Gaussian primitives, compressing the scene representation without sacrificing photorealistic fidelity. Comprehensive evaluations confirm that our method significantly outperforms existing baselines in tracking precision, rendering detail, and memory economy, demonstrating exceptional scalability for ultra-long driving sequences. We hope this joint optimization of robust tracking and memory-efficient mapping can provide valuable insights for future research on scaling high-fidelity visual SLAM to large-scale environments.

\section*{Acknowledgements}
This work is supported by the National Natural Science Foundation of China (No. 62406267), Guangdong Provincial Project (No. 2024QN11X072), Guangzhou-HKUST(GZ) Joint Funding Program (No. 2025A03J3956) and Guangzhou Municipal Education Project (No. 2024312122).

% ===================== Supplementary Material =====================
\clearpage
\appendix

\begin{center}
{\Large\bfseries Supplementary Material for KiloGS-SLAM}\\[1mm]
% {\large Robust and Efficient Monocular 3D Gaussian SLAM for Kilometer-Scale Outdoor Scenes}
\end{center}

% \vspace{1em}

%=================================================================%
\section{Overview}
%=================================================================%
This supplementary material provides further technical details and additional results for KiloGS-SLAM. The contents are organized as follows:
Section \ref{sec:app_tracking} details the Motion-Adaptive Hybrid Tracking module, including the flow-based keyframe selection and dual-modal pose solver. 
Section \ref{sec:app_loop} describes the asynchronous loop closure and global map alignment mechanism.
Section \ref{sec:app_mapping} provides the mathematical formulations for the hierarchical map representation and optimization. 
Section \ref{sec:app_management} formulates the multi-view metrics used in the Lifecycle-Managed Gaussian Mapping.
Section \ref{sec:app_hyperparameters} lists the detailed system hyperparameters.
Section \ref{sec:app_limitations} discusses the current limitations and future work.
Finally, Section \ref{sec:app_results} presents additional qualitative visualizations.

%=================================================================%
\section{Details of Motion-Adaptive Hybrid Tracking}
\label{sec:app_tracking}
%=================================================================%

\subsection{Motion-Aware Keyframe Selection}
To maintain a well-conditioned tracking trajectory and avoid redundant computational overhead when the camera is stationary or moving slowly, we implement a lightweight, flow-based keyframe selection strategy. Crucially, this module acts as an efficient gatekeeper prior to deep feature extraction.

Instead of relying on heuristic spatial distance thresholds, we track sparse features from the last keyframe $\mathcal{K}$ to the current frame $t$ using the highly efficient Pyramidal Lucas-Kanade (PyrLK) optical flow. Let the corresponding 2D coordinates be $\mathbf{u}_{\mathcal{K}}^i$ and $\mathbf{u}_t^i$. We compute the displacement magnitude for each tracked feature as $m_i = \|\mathbf{u}_t^i - \mathbf{u}_{\mathcal{K}}^i\|_2$.

To prevent moving foreground objects (\eg, passing vehicles) from artificially inflating the perceived ego-motion, we evaluate the median flow magnitude across all tracked features rather than the mean. The current frame is selected as a new keyframe only if this median displacement exceeds a pixel threshold $\tau_{kf}$:
\begin{equation}
    \text{Median}(\{ m_1, m_2, \dots, m_N \}) > \tau_{kf},
\end{equation}
where we empirically set $\tau_{kf} = 24$. This $\mathcal{O}(N)$ operation strictly ensures that keyframes are inserted based on sufficient environmental parallax, significantly boosting the overall system efficiency.

\subsection{Dual-Modal Pose Estimator Details}
As introduced in the main text, relying on a single Essential matrix ($\mathcal{M}_E$) is fragile in outdoor environments due to geometric degeneracies (\eg, flat roads or pure rotations). To address this, our pose estimator dynamically routes the optimization between $\mathcal{M}_E$ and a planar Homography model ($\mathcal{M}_H$), governed by GRIC score. 

\noindent \textbf{Geometric Robust Information Criterion (GRIC).} The routing is governed by the GRIC score, which balances model fidelity against structural complexity. For a set of $N$ putative correspondences, let $e_i(\mathcal{M})$ denote the geometric residual of the $i$-th match. To avoid expensive non-linear reprojection error computation, we employ the first-order Sampson distance as an efficient and accurate approximation. The GRIC score for a given model is formulated as:
\begin{equation}
    \text{GRIC}(\mathcal{M}) = \sum_{i=1}^{N} \rho(e_i(\mathcal{M})) + \lambda_1 d_\mathcal{M} N + \lambda_2 k_\mathcal{M},
\end{equation}
where $d_\mathcal{M}$ represents the dimension of the underlying structure ($d_E = 3$ for general 3D scenes, $d_H = 2$ for coplanar scenes), and $k_\mathcal{M}$ is the degrees of freedom ($k_E = 5$, $k_H = 8$). The robust penalty function $\rho(e(\mathcal{M}))$ is defined as:
\begin{equation}
    \rho(e(\mathcal{M})) = \min \left( \frac{e^2}{\sigma^2}, 2(r - d_\mathcal{M}) \right),
\end{equation}
where $\sigma=0.8$ is the assumed standard deviation of measurement noise, and $r=4$ is the data dimension for 2D-2D point pairs. We empirically set the weighting parameters as $\lambda_1 = \ln(r)$ and $\lambda_2 = \ln(rN)$.

\noindent \textbf{Parallel Evaluation and Routing.} To ensure real-time efficiency and robustness, the model selection follows a parallel competitive pipeline. We first establish a baseline score $\text{GRIC}(\mathcal{M}_H)$ using standard RANSAC. Simultaneously, a thread pool executes multiple RANSAC trials to compute candidate Essential matrices $\mathcal{M}_E$, evaluating $\text{GRIC}(\mathcal{M}_E)$ for each. The pose estimation and switching process then proceeds as follows:

\begin{enumerate}
    \item \textbf{Degeneracy Check:} If $\text{GRIC}(\mathcal{M}_H) \le \text{GRIC}(\mathcal{M}_E)$, the system identifies a geometric degeneracy (\eg, a planar scene), and the current $\mathcal{M}_E$ is immediately discarded. Otherwise, we select the $\mathcal{M}_E$ with the maximum inliers to recover the 2D-2D relative pose.
    \item \textbf{Scale Recovery:} Since the recovered $\mathcal{M}_E$ pose inherently lacks scale, we compute the absolute scale by aligning the triangulated 3D points with the geometric prior (UniDepth).
    \item \textbf{PnP Fallback:} If the valid $\mathcal{M}_E$ yields a pure rotation (zero translation magnitude), or if the scale recovery fails, the 2D-2D tracking is deemed unreliable. The system decisively falls back to a Perspective-n-Point (PnP) solver. By re-projecting the reference frame's 2D features using its predicted depth into 3D points, the PnP solver directly estimates the pose against the current frame, seamlessly bypassing the scale ambiguity and remaining immune to rotational degeneracies.
\end{enumerate}

%=================================================================%
\section{Details of Loop Closure and Global Optimization}
\label{sec:app_loop}
%=================================================================%
To eliminate accumulated drift over kilometer-scale trajectories and ensure global geometric consistency, we incorporate a robust loop closure mechanism. This module operates asynchronously to the main tracking thread and corrects both camera poses and the Gaussian map representation.

\noindent \textbf{Loop Detection and Re-matching.} 
We utilize DBoW2 to retrieve visually similar candidate keyframes from the historical trajectory. To ensure high recall and precision, the feature extraction and indexing are performed concurrently. Once a loop candidate pair $(j, k)$ is identified, we perform re-matching between the current frame $k$ and the historical frame $j$ using our deep feature pipeline (DISK + LightGlue). This provides high-quality 2D-2D correspondences. We then utilize the Dual-Modal Pose Estimator (detailed in Section \ref{sec:app_tracking}) to robustly compute the relative transformation between the loop frames.

\noindent \textbf{Pose Graph Optimization.}
Due to the inherent scale ambiguity in monocular SLAM, cumulative scale drift is unavoidable over long distances. Therefore, we optimize the global trajectory in the $Sim(3)$ space rather than $SE(3)$ to simultaneously correct rotation, translation, and scale.
Let $S_i = (R_i, \mathbf{t}_i, s_i) \in Sim(3)$ denote the absolute similarity transformation of the $i$-th keyframe. The global pose graph is optimized by minimizing a joint cost function comprising sequential smoothness constraints and the newly detected loop closure constraints:
\begin{equation}
\begin{split}
    \min_{S_1, \dots, S_N} & \sum_{i=1}^{N-1} \left\| \log_{Sim(3)} \left( (\Delta S_{i,i+1})^{-1} \cdot S_i^{-1} \cdot S_{i+1} \right) \right\|_2^2 \\
    & + \sum_{(j,k) \in \mathcal{L}} \left\| \log_{Sim(3)} \left( (\Delta S_{j,k}^{loop}) \cdot S_j^{-1} \cdot S_k \right) \right\|_2^2,
\end{split}
\end{equation}
where $\Delta S_{i,i+1}$ represents the sequential relative transformation estimated by our tracking module, and $\Delta S_{j,k}^{loop}$ is the relative transformation computed from the loop closure re-matching. $\mathcal{L}$ denotes the set of all detected loop pairs. This non-linear least-squares problem is solved using the Levenberg-Marquardt algorithm.

\noindent \textbf{Global Map Alignment.}
Once the globally optimal trajectory is recovered, the explicit Gaussian map must be updated accordingly to maintain spatial coherence. For each hierarchical anchor $\mathbf{p}_i \in \mathbb{R}^3$ originally spawned at the $m$-th keyframe, we apply a rigid transformation using its old pose $T_{old}^{(m)}$ and the optimized new pose $T_{new}^{(m)}$:
\begin{equation}
    \mathbf{p}_i^{new} = T_{new}^{(m)} \cdot (T_{old}^{(m)})^{-1} \cdot \mathbf{p}_i.
\end{equation}
To prevent excessive memory spikes during this global update, the transformation is executed in parallelized batches. After the spatial alignment, the updated anchors undergo a rapid re-voxelization and spatial hashing process (as described in Section \ref{sec:app_mapping}) to merge overlapping regions and maintain a compact, unified map structure.

%=================================================================%
\section{Details of Map Representation and Optimization}
\label{sec:app_mapping}
%=================================================================%
This section provides detailed mathematical formulations for the hierarchical map representation introduced in Section 3.2 of the main text.

\subsection{Voxelized Representation and MLP Decoding}
As outlined in the main text, the scene is parameterized by a set of sparse anchors $\mathcal{A} = \{\mathbf{p}_i \in \mathbb{R}^3\}$. Following Scaffold-GS, each anchor $i$ stores a neural context feature $\mathbf{f}_i$, a local scaling factor $\mathbf{l}_i \in \mathbb{R}^3$, and $K$ learnable positional offsets $\mathbf{O}_i \in \mathbb{R}^{K \times 3}$. For a given camera viewpoint $v$ with center $\mathbf{c}_v$, we compute the distance $d_{i,v} = \|\mathbf{p}_i - \mathbf{c}_v\|_2$ and the unit viewing direction $\mathbf{d}_{i,v} = (\mathbf{p}_i - \mathbf{c}_v) / d_{i,v}$.

To generate $K$ explicit 3D Gaussian primitives per anchor for rasterization, lightweight MLPs dynamically decode the anchor's neural feature $\mathbf{f}_i$, conditioned on $d_{i,v}$ and $\mathbf{d}_{i,v}$:
\begin{equation}
    (\mathbf{c}_{i,k}, \alpha_{i,k}, \mathbf{q}_{i,k}, \Delta \mathbf{s}_{i,k}, \Delta \mathbf{O}_{i,k}) = \text{MLP}(\mathbf{f}_i, \mathbf{d}_{i,v}, d_{i,v}), \quad k \in \{1, \dots, K\}.
\end{equation}
Crucially, the final center position of the $k$-th Gaussian primitive in world coordinates is scaled by the anchor's local scale $\mathbf{l}_i$:
\begin{equation}
    \mathbf{x}_{i,k} = \mathbf{p}_i + (\mathbf{O}_{i,k} + \Delta \mathbf{O}_{i,k}) \odot \mathbf{l}_i,
\end{equation}
where $\odot$ denotes element-wise multiplication. Similarly, its 3D scaling vector is modulated as $\mathbf{s}_{i,k} = \mathbf{l}_i \odot \Delta \mathbf{s}_{i,k}$.

\subsection{Depth Volume Rendering}
To enforce geometric constraints during optimization, we render the expected depth $\hat{D}$ alongside color. For a pixel $\mathbf{u}$, the rendered depth is computed via alpha compositing of the $z$-coordinates of the sorted Gaussians along the ray:
\begin{equation}
    \hat{D}(\mathbf{u}) = \sum_{j=1}^{\mathcal{N}} T_j \alpha_j z_j,
\end{equation}
where $T_j = \prod_{m=1}^{j-1} (1 - \alpha_m)$ denotes the accumulated transmittance of the ray reaching the $j$-th Gaussian primitive.

\subsection{Level-of-Detail (LoD) Mask Computation}
To prevent excessive fine-grained rendering in distant regions, we apply a distance-aware LoD activation mask. For an anchor $\mathbf{p}_i$ at resolution level $l$ (with voxel size $\epsilon_l$), its binary mask $\mathcal{M}_l(d_{i,v})$ is determined by its distance to the camera: it equals 1 within the optimal rendering range $[D_{min}^l, D_{max}^l]$, and 0 otherwise. 
The decoded opacity is modulated by this mask prior to rasterization:
\begin{equation}
    \tilde{\alpha}_{i,k} = \alpha_{i,k} \cdot \mathcal{M}_l(d_{i,v}).
\end{equation}
This mechanism automatically culls fine-grained Gaussians far from the camera, strictly bounding the memory and rendering overhead without sacrificing visual fidelity.

\subsection{Rendering Objectives}
During the sliding-window optimization, the total loss $\mathcal{L}_{total}$ comprises three components. 

First, the Photometric Loss ($\mathcal{L}_{photo}$) combines the $L_1$ distance and the D-SSIM term:
\begin{equation}
    \mathcal{L}_{photo} = 0.8 \|I - \hat{I}\|_1 + 0.2 (1 - \text{SSIM}(I, \hat{I})).
\end{equation}

Second, the Geometric Loss ($\mathcal{L}_{geo}$) penalizes the $L_1$ discrepancy between the rendered depth $\hat{D}$ and the predicted metric depth $D$:
\begin{equation}
    \mathcal{L}_{geo} = \|D - \hat{D}\|_1.
\end{equation}

Finally, since the unconstrained rasterization of 3DGS may cause primitives to stretch excessively along the viewing ray (creating severe artifacts during tracking), we incorporate an Isotropic Regularization ($\mathcal{L}_{iso}$) inspired by MonoGS. It penalizes the scaling parameters $\mathbf{s}_i$ of each Gaussian by their difference to their mean value $\tilde{s}_i$:
\begin{equation}
    \mathcal{L}_{iso} = \sum_{i=1}^{|\mathcal{G}|} \|\mathbf{s}_i - \tilde{s}_i \cdot \mathbf{1}\|_1.
\end{equation}

\subsection{Dynamic Map Expansion}
When integrating new keyframes or triggering densification via our Gaussian Lifecycle Management, the map expands dynamically. For any newly initialized 3D point $\mathbf{x}$, we determine its target LoD level $l$ based on its camera distance and compute its discrete voxel index $\mathbf{idx} = \lfloor \mathbf{x} / \epsilon_l \rfloor$. We utilize a Spatial Hash Table for $\mathcal{O}(1)$ state queries. If the voxel index is unoccupied, a new anchor is instantiated at the voxel center $\mathbf{p} = \mathbf{idx} \cdot \epsilon_l$. The neural feature $\mathbf{f}$ is zero-initialized, while the base scale $\mathbf{I}$ and offsets $\mathbf{O}$ are initialized using local geometric priors. This hashing mechanism ensures unbounded and memory-efficient map expansion.

%=================================================================%
\section{Details for Lifecycle-Managed Gaussian Mapping}
\label{sec:app_management}
%=================================================================%
In Section 3.3 of the main text, we introduced a scalable chunk-based mechanism to evaluate the densification score $S_{d}^{(i, c)}$ and the pruning score $S_{p}^{(i, c)}$ for each anchor $i$ within a local chunk $c$. This section provides the detailed mathematical formulations for these multi-view error-based metrics.

Given a chunk $c$, we uniformly sample a subset of viewpoints $V_c$. For each sampled view $v \in V_c$, let $I_v$ and $\hat{I}_v$ denote the ground-truth and rendered images, respectively. During the rasterization of view $v$, we track the pixel footprint of each anchor $i$, defining $\Omega_v(i)$ as the set of pixels where its decoded Gaussians actively contribute to the final color synthesis (\ie, their accumulated transmittance is above a minimal threshold).

\subsection{Densification Score ($S_{d}^{(i, c)}$)}
The densification score quantifies the frequency at which an anchor participates in synthesizing high-error regions, thereby identifying areas suffering from geometric under-reconstruction.

First, we compute a binary error mask $M_v(\mathbf{u})$ for each pixel $\mathbf{u}$ in view $v$, isolating regions where the photometric error exceeds a threshold $\tau_{err}$:
\begin{equation}
    M_v(\mathbf{u}) = 
    \begin{cases} 
        1, & \text{if } \| I_v(\mathbf{u}) - \hat{I}_v(\mathbf{u}) \|_1 > \tau_{err}, \\ 
        0, & \text{otherwise}.
    \end{cases}
\end{equation}
Next, we count the number of high-error pixels from anchor $i$ in view $v$:
\begin{equation}
    \kappa_v(i) = \sum_{\mathbf{u} \in \Omega_v(i)} M_v(\mathbf{u}).
\end{equation}
The chunk-level densification score $S_{d}^{(i, c)}$ is the average high-error pixel count across the sampled multi-view subset $V_c$:
\begin{equation}
    S_{d}^{(i, c)} = \frac{1}{|V_c|} \sum_{v \in V_c} \kappa_v(i).
\end{equation}
An anchor with a consistently high $S_{d}^{(i, c)}$ indicates that the current local geometry is insufficient to represent the high-frequency details across multiple views. As formulated in the main text, this triggers the global spawning of new anchors via a logical OR operation across all chunks.

\subsection{Pruning Score ($S_{p}^{(i, c)}$)}
The pruning score evaluates an anchor's overall contribution to the degradation of rendering quality. We first calculate the raw photometric contribution $\gamma_v(i)$ of anchor $i$ in view $v$, defined as the product of the view's overall photometric loss $\mathcal{L}_{photo}^{(v)}$ and the anchor's active pixel coverage $|\Omega_v(i)|$:
\begin{equation}
    \gamma_v(i) = \mathcal{L}_{photo}^{(v)} \cdot |\Omega_v(i)|.
\end{equation}
The cumulative raw degradation score for anchor $i$ across the chunk is:
\begin{equation}
    \tilde{S}_{p}^{(i, c)} = \sum_{v \in V_c} \gamma_v(i).
\end{equation}
In outdoor environments, the dynamic range of photometric losses can be extreme due to exposure variations or dynamic objects. To prevent the pruning metric from being dominated by extreme outliers and to ensure numerical stability, we apply a robust log-transformation:
\begin{equation}
    L^{(i, c)} = \log(1 + \tilde{S}_{p}^{(i, c)}).
\end{equation}
Finally, the score is min-max normalized across all active anchors $j \in \mathcal{A}_c$ within the chunk:
\begin{equation}
    S_{p}^{(i, c)} = \frac{L^{(i, c)} - \min_{j \in \mathcal{A}_c} L^{(j, c)}}{\max_{j \in \mathcal{A}_c} L^{(j, c)} - \min_{j \in \mathcal{A}_c} L^{(j, c)}}.
\end{equation}
This normalized score $S_{p}^{(i, c)} \in [0, 1]$ reliably isolates geometric artifacts and redundant floaters that persistently degrade view synthesis. Global pruning is then safely executed by applying max pooling across all chunks, as defined in the main text.

%=================================================================%
\section{Hyperparameter Settings}
\label{sec:app_hyperparameters}
%=================================================================%
We detail the primary hyperparameters used in our system across the Tracking, Map Representation and Optimization, and Lifecycle-Managed Gaussian Mapping modules in Table \ref{tab:hyperparameters}. The system is robust to these hyperparameters, and we keep them fixed across all datasets unless otherwise specified.

\begin{table}[h]
\centering
\fontsize{8pt}{9.5pt}\selectfont
\renewcommand{\arraystretch}{1.3}
\setlength{\tabcolsep}{4pt}
\caption{\textbf{System Hyperparameters.} The configuration of thresholds and weights used in KiloGS-SLAM.}
% \vspace{-5pt}
\begin{tabular}{clc} % 将第一列设为居中对齐 (c)
\toprule
\textbf{Module} & \textbf{Parameter Description} & \textbf{Value} \\
\midrule
\multirow{5}{*}{\textbf{Tracking}} 
& Keyframe selection flow threshold ($\tau_{kf}$) & 24 \\
& Motion prior angular threshold ($\tau_{R}$) & $5^\circ$ \\
& Motion prior translation angle ($\tau_{\theta}$) & $15^\circ$ \\
& GRIC assumed noise standard deviation ($\sigma$) & 0.8 \\
& Parallel RANSAC maximum trials & 20 \\
% & Cheirality check inlier ratio threshold & 10\% \\
\midrule
% 使用内嵌的 tabular 来实现居中换行
\multirow{8}{*}{\begin{tabular}{c}\textbf{Map Representation} \\ \textbf{and Optimization}\end{tabular}} 
& Number of LoD levels ($m$) &  5 \\
& voxel size ($\epsilon_1$) & 0.1  \\
& voxel size ($\epsilon_2$) & 0.25  \\
& voxel size ($\epsilon_3$) & 0.1  \\
& voxel size ($\epsilon_4$) & 5  \\
& voxel size ($\epsilon_5$) & 25  \\
% & Photometric loss $L_1$ weight & 0.8 \\
% & Photometric loss D-SSIM weight & 0.2 \\
& Geometric loss weight ($\lambda_g$) &  0.01 \\
& Isotropic loss weight ($\lambda_{iso}$) &  10 \\
\midrule
% 使用内嵌的 tabular 来实现居中换行
\multirow{5}{*}{\begin{tabular}{c}\textbf{Lifecycle-Managed} \\ \textbf{Gaussian Mapping}\end{tabular}} 
& DoG-Depth weight balance ($\lambda$) & 0.9 \\
& Chunk size ($C$) & 100 \\
& Densification photometric error threshold ($\tau_{err}$) & 0.1 \\
& Densification score threshold ($\tau_d$) & 5 \\
& Pruning score threshold ($\tau_p$) & 0.99 \\
\bottomrule
\label{tab:hyperparameters}
\end{tabular}
\end{table}

%=================================================================%
\section{Limitations and Future Work}
\label{sec:app_limitations}
%=================================================================%
While KiloGS-SLAM achieves state-of-the-art performance in tracking robustness and memory-efficient mapping for kilometer-scale scenes, it possesses certain limitations that provide avenues for future research.

First, although our efficient map representation and Gaussian lifecycle management successfully bound the computational and memory costs of long-sequence 3DGS-SLAM to an acceptable level, strict real-time performance remains a challenge. Recently, emerging approaches have begun exploring end-to-end streaming autoregressive networks (e.g., LongStream) to directly infer camera poses and point clouds across thousands of frames in outdoor scenes. Developing a streaming, autoregressive feed-forward Gaussian SLAM framework represents a highly promising direction to achieve true real-time capabilities.

Second, while our dynamic filtering mechanism mitigates the impact of moving objects during tracking, and our multi-view consistency-based pruning eliminates the resulting floaters, our current framework lacks explicit modeling of dynamic objects. Given that outdoor environments frequently feature moving vehicles and pedestrians, integrating our pipeline with 4D Gaussian representations to reconstruct large-scale dynamic scenes would significantly enhance its real-world applicability. This remains a primary objective for our future work.

%=================================================================%
% --- Tracking Figure Placeholder ---
\begin{figure*}[h]
    \centering
    \includegraphics[width=\textwidth]{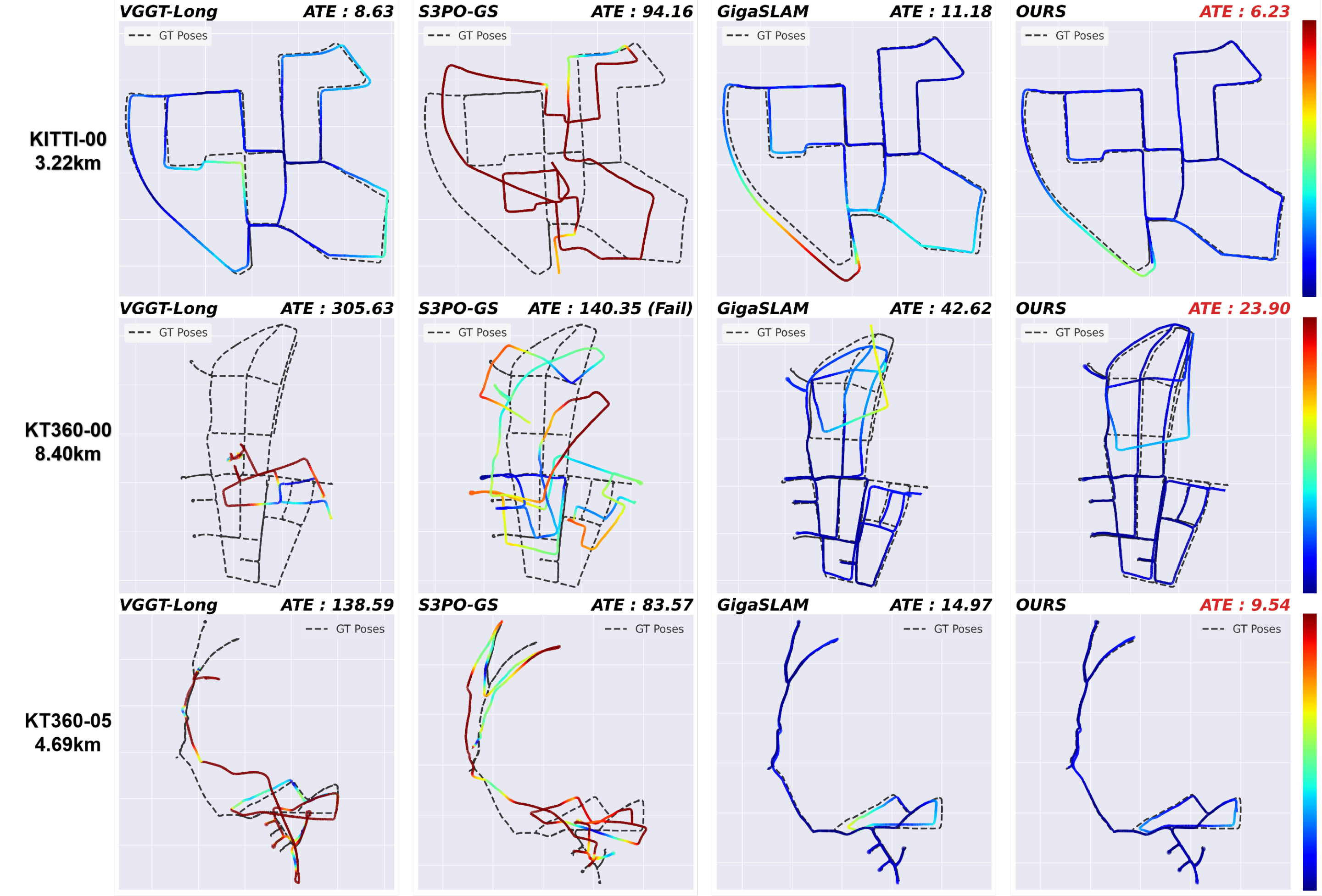}
    % \framebox[\textwidth]{\rule{0pt}{150pt} Tracking Trajectory Placeholder}
    \caption{\textbf{Additional tracking trajectories.} Qualitative evaluation of camera poses on extended outdoor sequences. KiloGS-SLAM reliably mitigates drift and maintains strong global consistency compared to other approaches.}
    \label{fig:app_tracking}
\end{figure*}

\section{Additional Qualitative Results}

To further validate the robustness and visual fidelity of KiloGS-SLAM, we provide additional qualitative results in this section.

\textbf{Tracking Robustness.} Figure \ref{fig:app_tracking} visualizes the estimated trajectories compared to the ground truth and baseline methods on extended sequences. Our system consistently maintains global consistency without catastrophic drift.

\begin{figure*}[h]
    \centering
    \includegraphics[width=\textwidth]{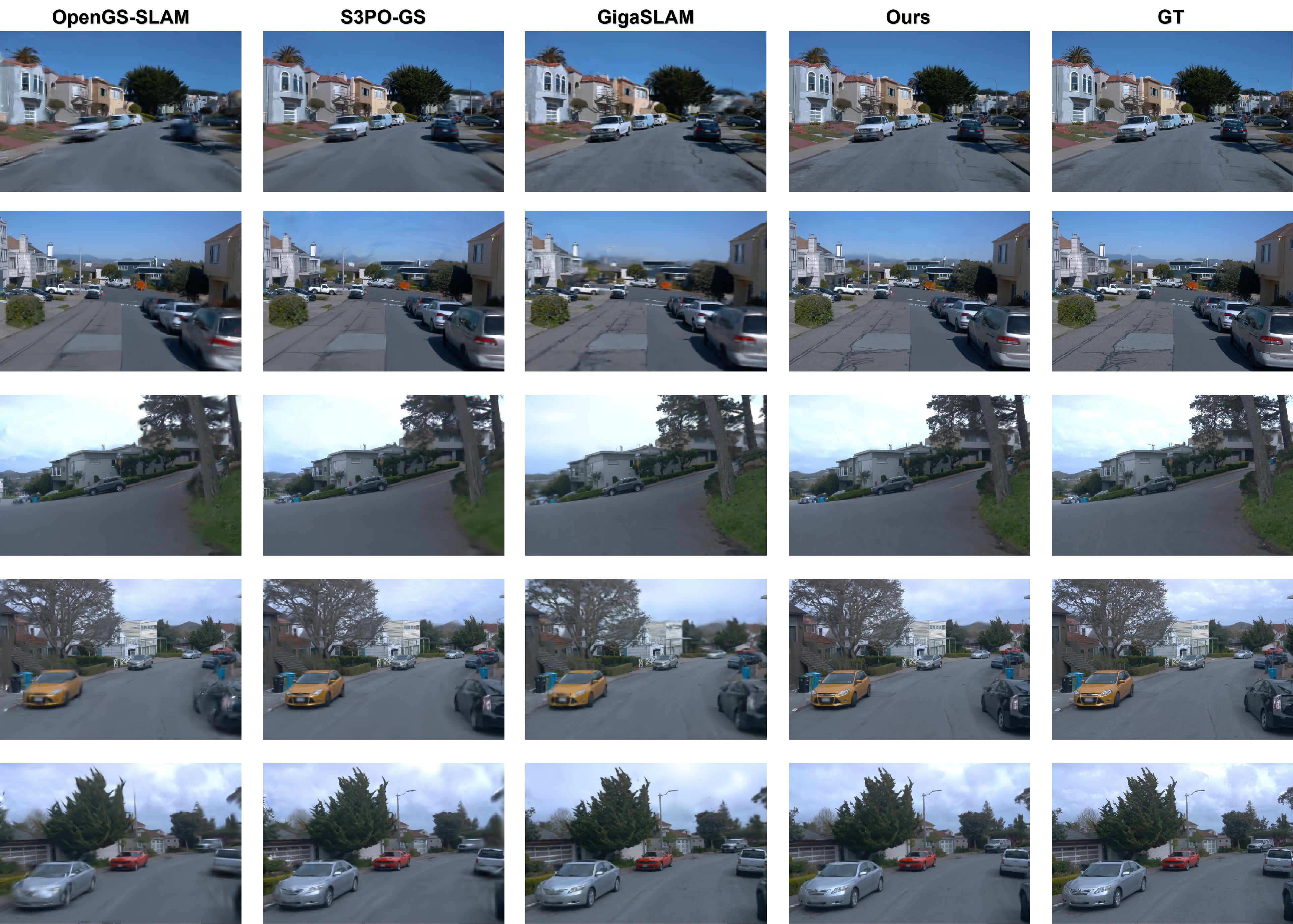}
    % \framebox[\textwidth]{\rule{0pt}{200pt} Waymo Rendering Comparisons Placeholder}
    \caption{\textbf{Additional qualitative rendering results on the Waymo dataset.} KiloGS-SLAM accurately synthesizes both complex near-field foregrounds and distant background structures with minimal artifacts.}
    \label{fig:app_render_waymo}
\end{figure*}

\begin{figure*}[t]
    \centering
    \includegraphics[width=\textwidth]{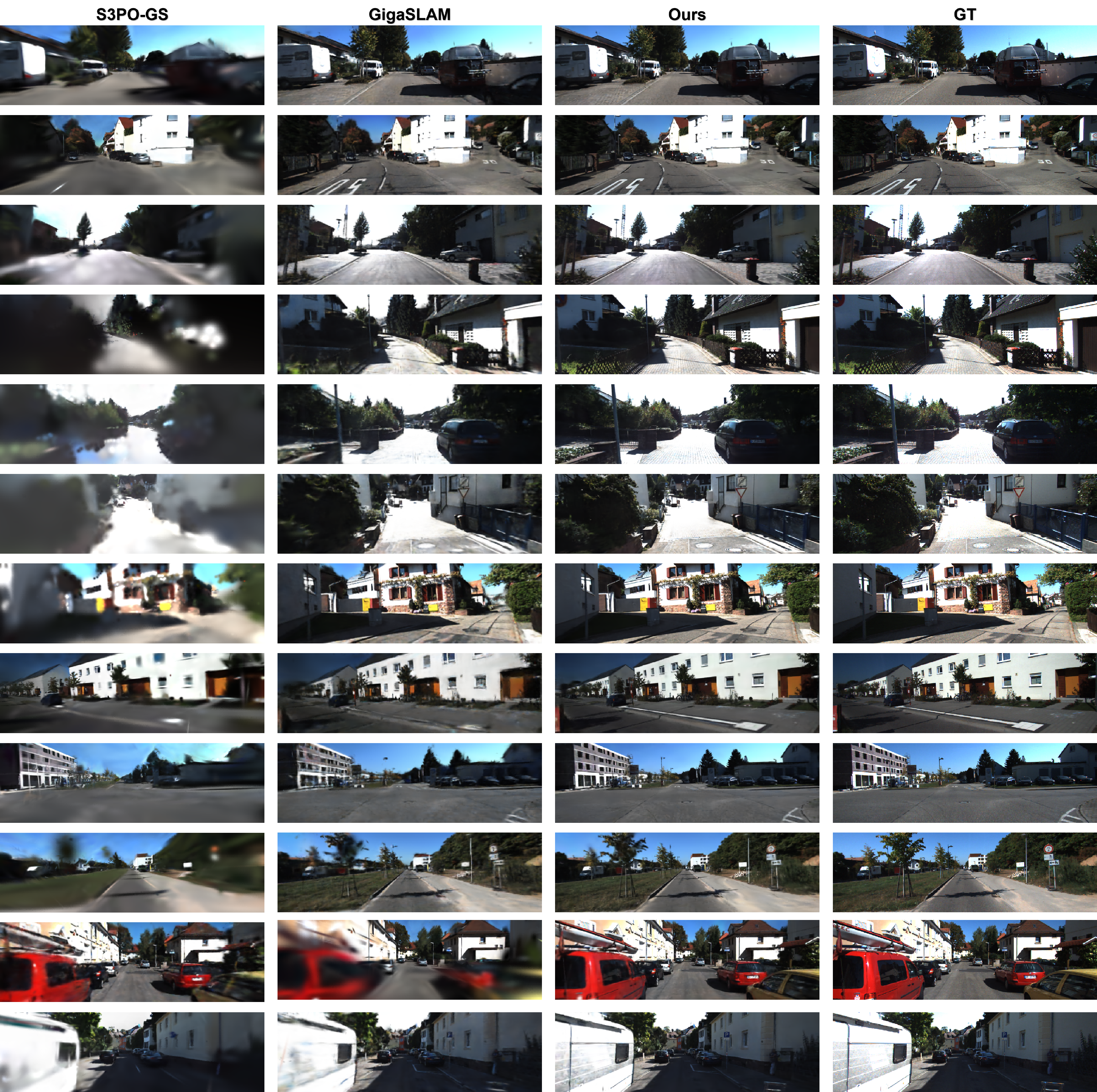}
    % \framebox[\textwidth]{\rule{0pt}{200pt} KITTI Rendering Comparisons Placeholder}
    \caption{\textbf{Additional qualitative rendering results on the KITTI dataset.} Compared to baselines that suffer from over-smoothing or aggressive pruning, our method retains high-frequency textural details (\eg, foliage, road markings).}
    \label{fig:app_render_kitti}
\end{figure*}

\textbf{Rendering Quality.} Figure \ref{fig:app_render_waymo} and Figure \ref{fig:app_render_kitti} showcase novel view synthesis comparisons on the Waymo and KITTI datasets, respectively. Thanks to our lifecycle-managed mapping, KiloGS-SLAM effectively suppresses floaters and preserves high-frequency details, yielding noticeably sharper boundaries and textures than the baselines.

\label{sec:app_results}
%=================================================================%
% --- Rendering Figure (Waymo) Placeholder ---

% --- Rendering Figure (KITTI) Placeholder ---

\clearpage

% ---- Bibliography ----
%
% BibTeX users should specify bibliography style 'splncs04'.
% References will then be sorted and formatted in the correct style.
%
\bibliographystyle{splncs04}
\bibliography{main}
\end{document}